\documentclass[11pt]{article}

\usepackage[preprint]{acl}

\usepackage{times}
\usepackage{latexsym}

\usepackage[T1]{fontenc}

\usepackage[utf8]{inputenc}

\usepackage{microtype}

\usepackage{inconsolata}

\usepackage{graphicx}
\usepackage{multirow}
\usepackage{booktabs}
\usepackage{makecell}
\usepackage[table]{xcolor}
\usepackage{enumitem}
\setitemize[1]{itemsep=-1pt,partopsep=0pt,topsep=1pt}
\usepackage{amsmath}
\usepackage{mathtools}
\usepackage{amssymb}
\usepackage{tcolorbox}

\title{\raisebox{-0.1\height}{\includegraphics[height=1.0em]{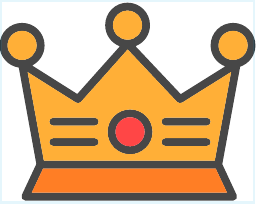}} DiaDem: Advancing \underline{Dia}logue \underline{De}scriptions in Audiovisual Video Captioning for Multi\underline{m}odal Large Language Models}

\author{Xinlong Chen\textsuperscript{\textmd{1,2,3}}\thanks{This work was conducted during the author's internship at Kling Team, Kuaishou Technology},
Weihong Lin\textsuperscript{\textmd{3}},
Jingyun Hua\textsuperscript{\textmd{3}},
Linli Yao\textsuperscript{\textmd{4}}, \\
\textbf{Yue Ding}\textsuperscript{\textmd{1,2}},
\textbf{Bozhou Li}\textsuperscript{\textmd{4}},
\textbf{Bohan Zeng}\textsuperscript{\textmd{4}},
\textbf{Yang Shi}\textsuperscript{\textmd{4}}, \\
\textbf{Qiang Liu}\textsuperscript{\textmd{1,2}\thanks{Corresponding author: qiang.liu@nlpr.ia.ac.cn}},
\textbf{Yuanxing Zhang}\textsuperscript{\textmd{3}},
\textbf{Pengfei Wan}\textsuperscript{\textmd{3}},
\textbf{Liang Wang}\textsuperscript{\textmd{1,2}},
\textbf{Tieniu Tan}\textsuperscript{\textmd{1,2,5}} \\
\textsuperscript{\textmd{1}}New Laboratory of Pattern Recognition, Institute of Automation, Chinese Academy of Sciences \\
\textsuperscript{\textmd{2}}School of Artificial Intelligence, University of Chinese Academy of Sciences \\ 
\textsuperscript{\textmd{3}}Kling Team, Kuaishou Technology \quad
\textsuperscript{\textmd{4}}Peking University \quad
\textsuperscript{\textmd{5}}Nanjing University \\[4pt]
\textbf{\texttt{Project webpage:~\url{https://diadem-captioner.github.io/}}}
}

\begin{document}
\maketitle
\begin{abstract}
Accurate dialogue description in audiovisual video captioning is crucial for downstream understanding and generation tasks. However,
existing models generally struggle to produce faithful dialogue descriptions within audiovisual captions. To mitigate this limitation, we propose \textbf{DiaDem}, a powerful audiovisual video captioning model capable of generating captions with more precise dialogue descriptions while maintaining strong overall performance. We first synthesize a high-quality dataset for SFT, then employ a difficulty-partitioned two-stage GRPO strategy to further enhance dialogue descriptions. To enable systematic evaluation of dialogue description capabilities, we introduce \textbf{DiaDemBench}, a comprehensive benchmark designed to evaluate models across diverse dialogue scenarios, emphasizing both speaker attribution accuracy and utterance transcription fidelity in audiovisual captions. Extensive experiments on DiaDemBench reveal even commercial models still exhibit substantial room for improvement in dialogue-aware captioning. Notably, DiaDem not only outperforms the Gemini series in dialogue description accuracy but also achieves competitive performance on general audiovisual captioning benchmarks, demonstrating its overall effectiveness.
\end{abstract}

\section{Introduction}
Audiovisual joint video captioning has recently garnered increasing attention~\citep{tang2025video, wu2025ugc}. Compared to visual-only video captioning~\citep{wang2024tarsier, chen2025vidcapbench, yuan2025tarsier2, chai2024auroracap, chen2025versavid, zhong2025owlcap, shi2025mavors}, incorporating auditory signals enables models to generate more comprehensive and human-aligned textual descriptions~\citep{chen2025avocado, ma2025omni}. High-quality audiovisual captions not only facilitate effective alignment of audio, visual, and textual modalities during pretraining~\citep{zhang2024video, wang2025haic}, but also provide critical support for downstream audiovisual understanding and generation tasks~\citep{du2025vc4vg, ren2025anycap, hua2025vabench, ge2025framemind, shi2025realunify}.

\begin{figure*}[t]
  \includegraphics[width=\linewidth]{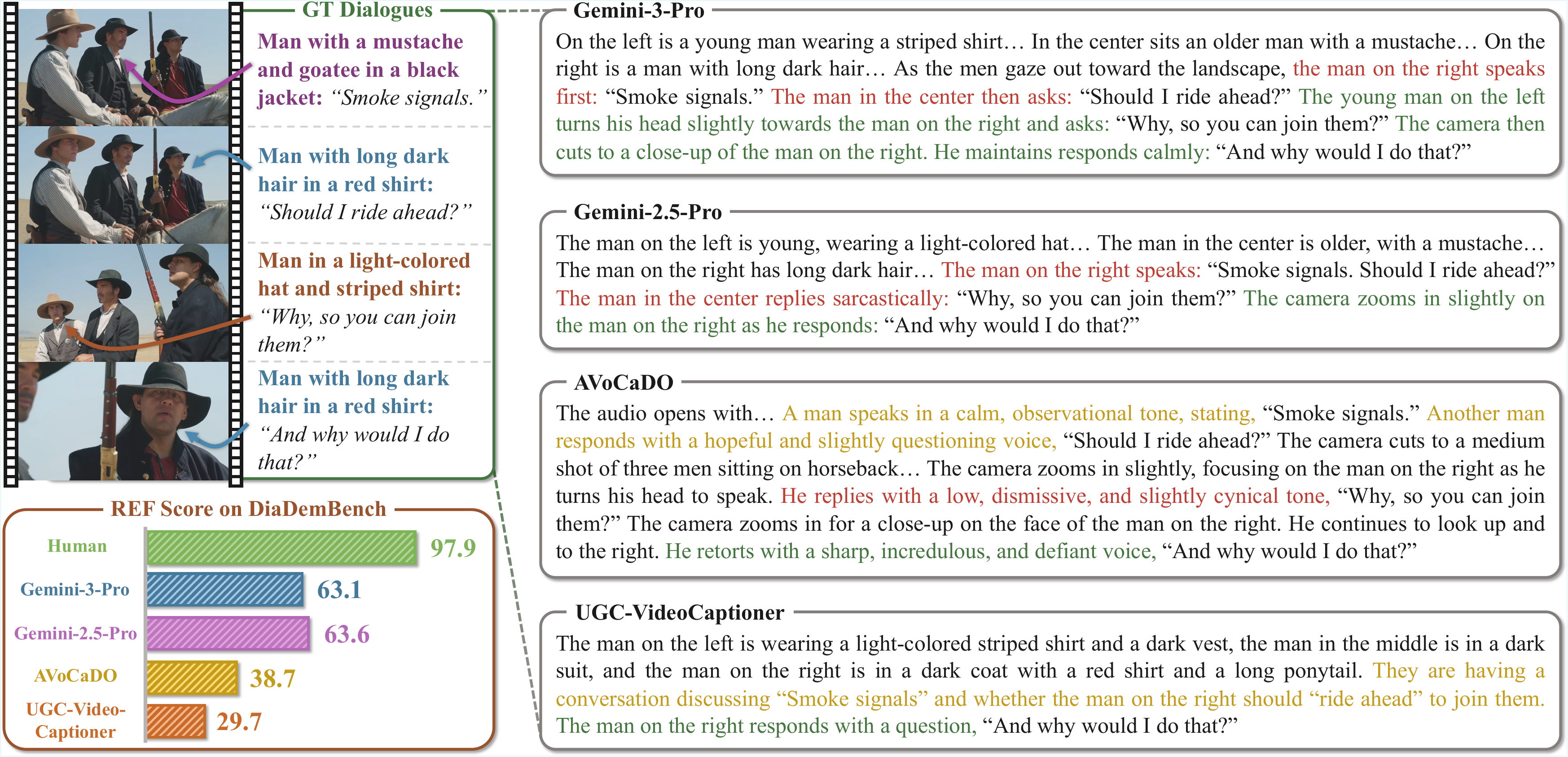}
  \caption{Qualitative and quantitative analysis of dialogue description capabilities in existing models, highlighting substantial room for improvement in speaker-utterance attribution. In the generated audiovisual captions, speaker attributions are color-coded as \textcolor[HTML]{ba271c}{incorrect}, \textcolor[HTML]{be9002}{ambiguous}, and \textcolor[HTML]{2e672b}{correct}.}
  \label{fig:teaser}
\end{figure*}

Despite notable progress in audiovisual video captioning, most existing models and benchmarks primarily prioritize descriptive completeness~\citep{tang2025video, wu2025ugc}, while overlooking a critical dimension in audiovisual scenarios: \textit{the accuracy of dialogue description, encompassing precise utterance transcription and correct speaker attribution}. As illustrated in Fig.~\ref{fig:teaser}, current models consistently struggle to discern ``who said what'' in challenging multi-party dialogue scenarios. However, faithful speaker-utterance identification and attribution in audiovisual captioning are indispensable for accurately capturing characters' intentions, inferring interpersonal dynamics, and reconstructing the narrative's logical structure.

To enable systematic and fine-grained evaluation of dialogue description capabilities in audiovisual captioning, we first introduce \textbf{DiaDemBench}, which comprises 1,039 manually annotated videos featuring diverse dialogue settings, covering single- and multi-shot scenes, varying speaker counts, and challenging scenarios with overlapping speech, thereby providing a comprehensive testbed for evaluating both utterance transcription fidelity and speaker attribution accuracy in audiovisual captioning. As shown in Fig.~\ref{fig:teaser}, even advanced open-source and commercial models still exhibit substantial room for improvement in speaker-utterance matching, with a notable performance gap persisting between open-source and commercial models.

To address this issue, we further propose \textbf{DiaDem}, an audiovisual video captioning model capable of more precise utterance transcription and speaker attribution, while preserving the completeness of other audiovisual details. Building upon AVoCaDO~\citep{chen2025avocado}, which possesses strong capability in holistic audiovisual captioning, DiaDem is enhanced through targeted post-training to improve dialogue understanding. Specifically, leveraging the complementary strengths of different models, we design a dedicated pipeline to construct a training corpus comprising 70K high-quality audiovisual captions with precise dialogue descriptions, together with 15K general-purpose captions from non-dialogue scenes for SFT, equipping the model with foundational dialogue description skills while maintaining its general captioning performance. Furthermore, we manually annotate 3K dialogue-centric samples and incorporate them into a difficulty-partitioned two-stage GRPO strategy, further strengthening the model’s ability to transcribe utterances and attribute them to the correct speakers in audiovisual captions. Experimental results show that DiaDem not only outperforms the Gemini series in dialogue description accuracy, but also achieves competitive performance in general audiovisual captioning.

Our contributions are summarized as follows:
\begin{itemize}[leftmargin=*]
    \item We propose DiaDem, a powerful audiovisual video captioning model that delivers superior dialogue description capability compared to the Gemini series, while achieving competitive general audiovisual captioning quality.
    \item We introduce DiaDemBench, the first benchmark designed to evaluate the accuracy of dialogue descriptions in audiovisual models, covering both utterance transcription and speaker attribution.
    \item We design a post-training strategy that first synthesizes a high-quality dataset for SFT, followed by a difficulty-partitioned two-stage GRPO strategy to sharpen more precise dialogue description and boost the overall captioning performance.
\end{itemize}

\section{Related Works}
\subsection{\hspace{-5pt}Audiovisual \hspace{-0.5pt}Video \hspace{-0.5pt}Captioning \hspace{-0.5pt}with \hspace{-0.5pt}MLLMs}
With the rapid advancement of audiovisual understanding models~\citep{hou2024toward, cheng2024videollama, shu2025audio, panagopoulou2023x, ye2024cat, guo2025aligned, sun2024video, team2025longcat}, audiovisual captioning models and benchmarks have also evolved accordingly. For instance, video-SALMONN-2~\citep{tang2025video} introduces MrDPO to mitigate information loss and hallucinations, and proposes a testset that evaluates captions  based on the accuracy and completeness of atomic events. UGC-VideoCaptioner~\citep{wu2025ugc} enhances captioning performance by distilling knowledge from Gemini-2.5-Flash~\citep{comanici2025gemini} combined with reinforcement learning (RL), and further introduces UGC-VideoCap, a benchmark where a judge model scores captions across multiple dimensions. AVoCaDO~\citep{chen2025avocado} adopts a two-stage post-training strategy that explicitly emphasizes the temporal alignment of audiovisual events to improve caption fidelity. Qwen3-Omni-Captioner~\citep{ma2025omni} designs an agentic data generation and cleaning pipeline to construct high-quality caption corpora, and proposes Omni-Cloze, which evaluates caption quality via a cloze-style multiple-choice proxy task.

However, existing works primarily focus on the completeness of descriptive content, often neglecting the accuracy of dialogue descriptions in audiovisual scenarios, which is critical for downstream understanding and generation tasks. To bridge this gap, we propose DiaDem, a captioning model that generates audiovisual captions with dialogue accuracy surpassing that of the Gemini series. Additionally, we introduce DiaDemBench, a benchmark specifically designed to evaluate the accuracy of dialogue descriptions in audiovisual captions.

\subsection{Speaker Diarization and Recognition}
In the audio-only domain, traditional speaker diarization frameworks~\citep{medennikov2020target, chen20253d} typically consist of multiple components, including Voice Activity Detection (VAD) to identify speech segments, speaker embedding extraction to capture distinctive speaker characteristics, and clustering to group embeddings for speaker identification. Subsequent studies have further incorporated Automatic Speech Recognition (ASR) modules to obtain speaker identities along with their utterances~\citep{WangJoint, cornell2024one}. However, due to the accumulation of errors across modular pipelines, later works have explored post-processing with LLMs~\citep{park2024enhancing, efstathiadis2025llm} or adopted end-to-end approaches to mitigate error propagation~\citep{liang2023second, yin2025speakerlm}.

Nevertheless, in the absence of visual information, these methods generally produce only speaker IDs paired with transcribed speech, lacking the descriptive information about speakers that is essential in audiovisual captioning. In contrast, DiaDem integrates both audio and visual modalities to generate high-quality audiovisual captions with accurate speaker descriptions and utterance recognition.

\section{DiaDemBench}
\subsection{Overview}
DiaDemBench is designed to comprehensively evaluate the accuracy of dialogue descriptions in audiovisual video captions, focusing on two critical yet underrepresented dimensions in existing benchmarks: \textit{correct speaker attribution} and \textit{precise utterance transcription}. To enable quantitative assessment along these axes, we introduce a suite of tailored evaluation metrics. Moreover, we collect a representative set of dialogue-centric videos spanning diverse scenarios, together with manually curated annotations to ensure both coverage and fidelity in evaluation.

\subsection{Evaluation Protocol}\label{sec: Evaluation Protocol}
The accuracy of dialogue descriptions comprise two aspects: \textit{correct speaker attribution} and \textit{precise utterance transcription}. Since speaker attribution is meaningful only when the utterance is accurately transcribed, our evaluation protocol follows a two-stage approach. Specifically, we first align predicted and ground-truth dialogue tuples based on utterance matching and compute the utterance accuracy score, denoted as $\mathrm{ASR}$. We then assess speaker consistency within the successfully matched tuples to obtain the speaker reference accuracy score, denoted as $\mathrm{REF}$. The detailed procedure is described below.

Given the structural variability of dialogue descriptions in audiovisual captions, we first employ Gemini-2.5-Pro to extract and structure the dialogue list from each caption. Let $P = [p_1, p_2, \ldots, p_M]$ denote the predicted dialogue sequence and $G = [g_1, g_2, \ldots, g_N]$ denote the ground-truth dialogue sequence, where $M$ and $N$ are the respective sequence lengths. For any element $x_i \in P \cup G$, we denote its utterance as $u(x_i)$ and its speaker as $s(x_i)$.

To quantify the similarity between any pair of utterances $u(p_i)$ and $u(g_j)$, where $i \in [1, M]$ and $j \in [1, N]$, we use the normalized edit distance:
$$\text{Sim}\big(u(p_i), u(g_j)\big) = 1 - \frac{\text{ed}\big(u(p_i), u(g_j)\big)}{\max\big(\big|(u(p_i)\big|, \big|u(g_j)\big|\big)}$$
where $\mathrm{ed}(\cdot)$ denotes the standard Levenshtein edit distance\footnote{\url{https://en.wikipedia.org/wiki/Levenshtein_distance}} between two strings.

\begin{figure*}[t]
  \includegraphics[width=\linewidth]{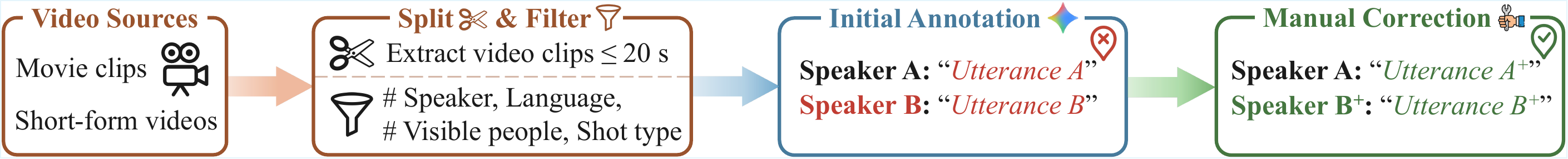}
  \caption{Overview of the curation pipeline of DiaDemBench.}
  \label{fig:bench_curation}
\end{figure*}

To achieve optimal matching between the two dialogue lists, we aim to find a subsequence of $P$ and a subsequence of $G$ with equal length such that (i) each matched utterance pair exceeds a predefined similarity threshold $\gamma$, and (ii) the sum of similarities across all matched pairs is maximized. This problem can be naturally formulated and solved via dynamic programming~\citep{bellman1966dynamic}.

\begin{figure}
  \includegraphics[width=\linewidth]{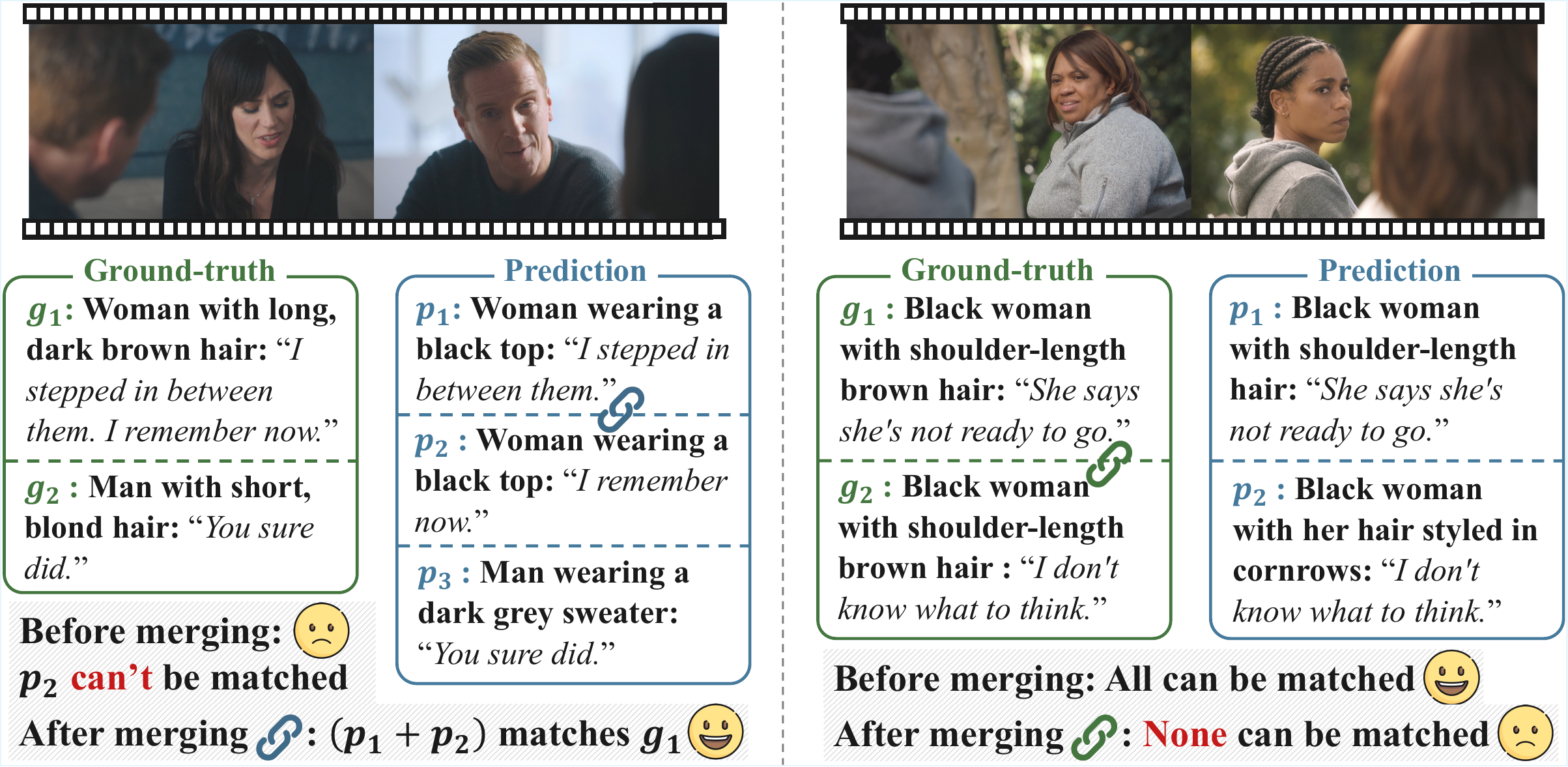}
  \caption{When the dialogue segmentation of the ground-truth annotations is inconsistent with that of the model predicts, issues arise regardless of whether adjacent utterances from the same speaker are merged.}
  \label{fig:dialogue_enhance}
\end{figure}

However, a naive one-to-one matching strategy, as used in AVoCaDO, fails to account for segmentation inconsistencies between predicted and ground-truth dialogues, which may lead to counterintuitive matching results. As illustrated in Fig.~\ref{fig:dialogue_enhance}, in the left example, naive matching causes $p_2$ to be omitted, whereas merging adjacent utterances from the same speaker resolves the issue. However, in the right example, naive matching aligns better with human intuition, while forced merging prevents both $p_1$ and $p_2$ from being successfully matched due to insufficient similarity. Moreover, the segmentation of consecutive utterances from the same speaker is inherently subjective and model-dependent, making it hard to define a canonical segmentation rule.

To mitigate the adverse effects of such segmentation variance, we propose an \textit{adaptive merging strategy}. Specifically, during dynamic programming, adjacent utterances from the same speaker are dynamically merged whenever doing so improves alignment fidelity, yielding matches that better reflect human intuition.

Formally, let $F_{i,j}$ denote the maximum cumulative utterance similarity between the first $i$ elements of $P$ and the first $j$ elements of $G$. The recurrence relation, which incorporates adaptive merging, is defined as follows:
$$
\hspace*{-4pt}
F_{i,j} = 
\begin{cases} 
0, \quad \text{if } i = 0 \text{ or } j = 0, \\
\max \big\{ F_{i-1,j}, F_{i,j-1} \big\}, \quad\ \ \text{if } i > 0,\, j > 0,\,\\\quad\quad\quad\quad\quad\quad\ \ \ \Phi\!\big(P_{i-k+1}^i,\, G_{j-l+1}^j\big) < \gamma \\\quad\quad\quad\quad\quad\quad\quad\quad\ \text{ for all } 1 \leq k,l \leq W, \\
\max \Big\{ F_{i-1,j}, F_{i,j-1}, \max\limits_{\substack{1 \leq k,\, l \leq W}} \big\{ F_{i-k, j-l} + \\\quad\quad\quad \Phi(P_{i-k+1}^i, G_{j-l+1}^j) \big\} \Big\}, \ \text{otherwise.} \\
\end{cases}$$
Here, $P_a^b$ and $G_a^b$ denote subsequences from index $a$ to $b$ in the respective dialogue sequences. $W$ controls the maximum merging window size, balancing matching accuracy against computational cost. $\Phi(\tilde{P}, \tilde{G})$ computes the utterance similarity between subsequences $\tilde{P}$ and $\tilde{G}$, defined as:

$$
\hspace*{-10pt}
\Phi(\tilde{P}, \tilde{G})=
\begin{cases}
-\infty, \text{if } \exists\, p_i, p_j \in \tilde{P} \text{ s.t. } s(p_i) \neq s(p_j) \\\quad\quad \text{ or } \exists\, g_i, g_j \in \tilde{G} \text{ s.t. } s(g_i) \neq s(g_j), \\
\text{Sim}\Big(\text{concat}_{\substack{p_i \in \tilde{P}}}u(p_i), \\\quad\quad\  \text{concat}_{\substack{g_j \in \tilde{G}}}u(g_j)\Big), \ \text{otherwise.}
\end{cases}$$

Importantly, merging is restricted to adjacent utterances from the same speaker to avoid introducing speaker ambiguity. Additional implementation details are provided in App.~\ref{app:asr_detail}.

\begin{figure*}[t]
  \includegraphics[width=\linewidth]{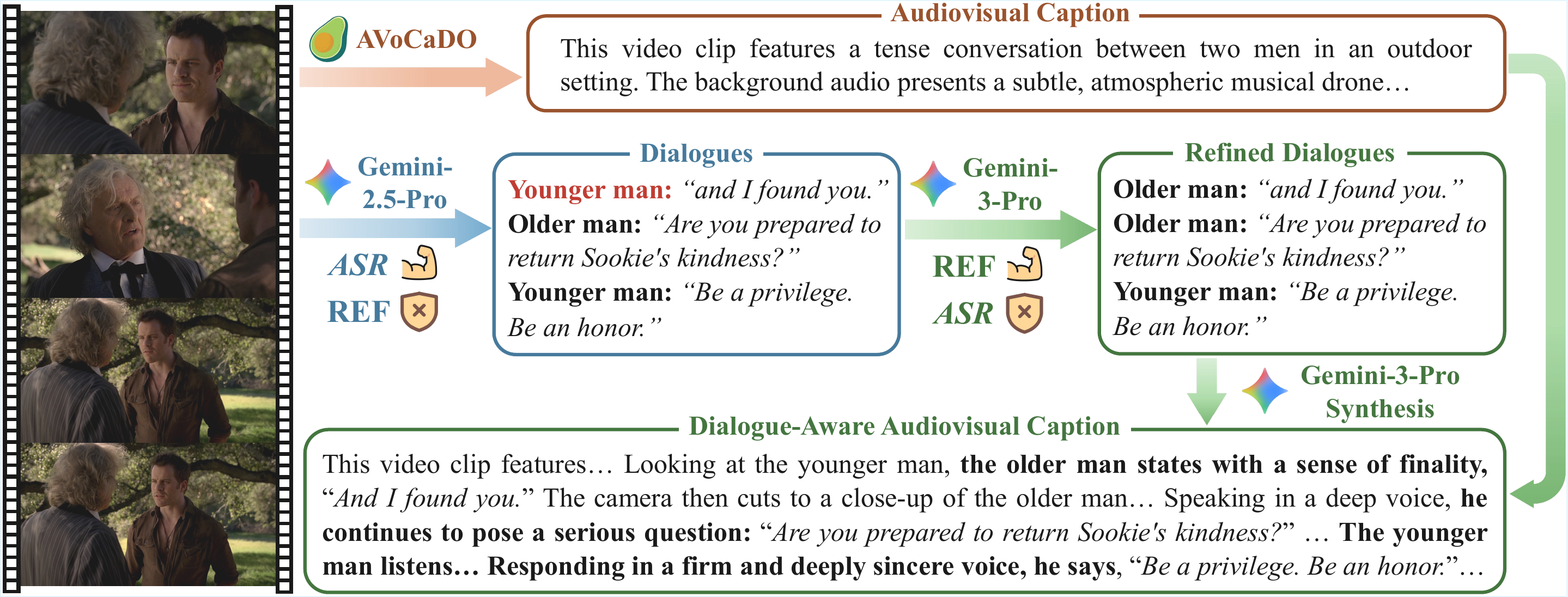}
  \caption{Data annotation pipeline for SFT.}
  \label{fig:sft_grpo_data_curation}
\end{figure*}

Upon completing utterance matching, we obtain the utterance accuracy score $S_{\text{utterance}} = F_{M,N}$ and the corresponding adaptively merged matched subsequences $P'$ and $G'$. We then compute the speaker reference accuracy score $S_{\text{speaker}}$ based on speaker consistency within each matched tuple. Specifically, we use Gemini-2.5-Flash as the judge model, which, conditioned on the video content $v$, assesses whether the speaker descriptions in each matched tuple are consistent and assigns a binary score:
$$S_{\text{speaker}}=\sum_{(p_i, g_i) \in (P', G')} \text{Judge}\big(s(p_i), s(g_i), v \big)$$
To ensure fair comparison across samples with varying dialogue lengths, both $S_{\text{speaker}}$ and $S_{\text{utterance}}$ are normalized. Let $M'$ and $N'$ denote the lengths of the adaptively merged sequences $P'$ and $G'$, respectively. The scores are bounded in $[0, \min(M', N')]$, enabling the computation of precision (normalized by $M'$), recall (normalized by $N'$), and the corresponding F1 score. These F1 scores serve as our final evaluation metrics:
$$\mathrm{REF} = \text{F1}(S_{\text{speaker}}, M', N')$$
$$\mathrm{ASR} = \text{F1}(S_{\text{utterance}}, M', N')$$

\subsection{Data Curation}
As illustrated in Fig.~\ref{fig:bench_curation}, we begin by collecting movie clips and short-form videos from public user‑generated content (UGC) platforms that feature rich dialogue scenarios. Considering the context window limitations of current models, raw videos are segmented into clips no longer than 20 seconds using PySceneDetect\footnote{\url{https://www.scenedetect.com/}}. We then employ multiple open-source models to infer and cross-validate key attributes of each clip, such as speaker counts, visible people counts, language information, shot editing types. Based on these attributes, we filter the clips and obtain 1,039 video segments with broad category coverage and balanced distributions. Each selected clip is subsequently annotated with fine-grained dialogue descriptions. To improve annotation efficiency, we first use Gemini-2.5-Pro to generate an initial dialogue description, which is then manually refined to ensure precise utterance transcriptions and correct speaker attribution. Details are provided in App.~\ref{app:anno_pipeline}.

\section{DiaDem}
\subsection{Overview}
After constructing DiaDemBench, we evaluate a wide range of models. The results in Tab.~\ref{tab:main result} indicate a notable performance gap persisting between open-source and commercial models. To mitigate this, we propose DiaDem, a powerful audiovisual video captioning model that integrates a carefully designed data annotation pipeline for SFT and a difficulty-partitioned two-stage GRPO strategy, thereby achieving superior dialogue description capability.

\begin{table*}
    \centering
    \resizebox{\linewidth}{!}{
    \begin{tabular}{l@{\hspace{-10pt}}c|>{\columncolor{gray!10}}c|c@{\hspace{8pt}}c@{\hspace{8pt}}c@{\hspace{8pt}}c@{\hspace{8pt}}c|c@{\hspace{8pt}}c@{\hspace{8pt}}c@{\hspace{8pt}}c@{\hspace{8pt}}c}
    \toprule
        \multirow{2}{*}[-0.1cm]{\textbf{Model}} & \multirow{2}{*}[-0.1cm]{\textbf{Size}} & \textbf{Overall} & \multicolumn{5}{c|}{\textbf{Single-shot} (REF / ASR)} & \multicolumn{5}{c}{\textbf{Multi-shot} (REF / ASR)} \\
        \cmidrule(lr){4-8} \cmidrule(lr){9-13}
        ~ & ~ & (REF / ASR) & \textbf{All} & $N=1$ & $N=2$ & $N{\ge}3$ & Overlap & \textbf{All} & $N=1$ & $N=2$ & $N{\ge}3$ & Overlap \\
        \midrule
        \textcolor{gray!100}{Human (avg. of 3 authors)} & \textcolor{gray!100}{-} & \textcolor{gray!100}{97.9 / 97.1} & \textcolor{gray!100}{98.0 / 97.4} & \textcolor{gray!100}{98.7 / 97.9} & \textcolor{gray!100}{98.1 / 97.5} & \textcolor{gray!100}{97.8 / 97.3} & \textcolor{gray!100}{87.3 / 89.2} & \textcolor{gray!100}{97.9 / 96.9} & \textcolor{gray!100}{98.5 / 97.4} & \textcolor{gray!100}{98.4 / 97.3} & \textcolor{gray!100}{98.1 / 97.2} & \textcolor{gray!100}{88.1 / 86.7} \\
        \midrule
        Gemini-2.5-Pro & - & \underline{63.6} / \underline{74.8} & \underline{54.2} / \underline{66.6} & \underline{62.3} / \underline{71.1} & \underline{59.4} / \underline{66.7} & \textbf{43.8} / \textbf{62.6} & 52.8 / \textbf{80.0} & 69.0 / \underline{79.6} & 64.5 / 77.3 & \textbf{75.6} / \underline{81.7} & \underline{67.8} / \underline{80.3} & \underline{55.4} / \textbf{68.0} \\
        Gemini-3-Pro & - & 63.1 / 71.0 & 51.8 / 59.8 & 61.2 / 62.0 & 56.6 / 61.1 & 40.5 / 56.5 & \underline{56.3} / 68.6 & \underline{69.7} / 77.5 & \underline{70.5} / \underline{80.7} & 71.6 / 77.0 & \textbf{68.5} / 76.8 & \textbf{57.2} / \underline{65.1} \\
        Gemini-2.5-Flash & - & 58.5 / 73.3 & 48.2 / 64.1 & 55.7 / 65.3 & 51.9 / 66.2 & 39.4 / 60.6 & \textbf{56.7} / \underline{77.8} & 64.3 / 78.6 & 64.5 / 78.4 & 68.3 / 78.4 & 62.3 / \underline{80.3} & 48.4 / 62.4 \\
        \midrule
        video-SALMONN-2 & 7B & 11.5 / 16.6 & 9.6 / 13.2 & 18.4 / 20.1 & 7.2 / 9.4 & 5.8 / 12.0 & 7.6 / 10.3 & 12.6 / 18.5 & 16.4 / 18.9 & 11.2 / 16.1 & 11.1 / 19.7 & 8.9 / 17.7 \\
        ARC-Qwen-Video & 7B & 11.7 / 17.2 & 8.2 / 12.4 & 16.1 / 15.3 & 7.0 / 13.2 & 4.0 / 9.9 & 0.0 / 5.5 & 13.4 / 19.6 & 17.5 / 25.0 & 15.3 / 20.4 & 9.3 / 15.8 & 0.0 / 2.8 \\
        HumanOmniV2 & 7B & 15.3 / 21.5 & 16.2 / 22.7 & 18.2 / 24.4 & 17.6 / 25.7 & 13.4 / 18.8 & 11.0 / 12.7 & 14.7 / 20.6 & 19.9 / 22.7 & 13.2 / 20.7 & 12.8 / 19.8 & 8.6 / 12.7 \\
        OmniVinci & 9B & 17.1 / 25.2 & 14.4 / 23.6 & 12.7 / 19.0 & 21.8 / 29.8 & 8.1 / 20.7 & 0.0 / 0.0 & 18.5 / 26.0 & 23.9 / 31.8 & 17.6 / 25.2 & 16.8 / 24.2 & 2.9 / 9.0 \\
        Qwen2.5-Omni & 7B & 26.1 / 37.1 & 26.4 / 36.8 & 38.6 / 48.8 & 24.6 / 33.0 & 18.9 / 31.5 & 10.2 / 28.6 & 25.9 / 37.1 & 35.0 / 45.7 & 27.7 / 37.3 & 18.4 / 32.1 & 11.3 / 14.6 \\
        UGC-VideoCaptioner & 3B & 29.7 / 47.0 & 31.0 / 44.3 & 50.8 / 56.7 & 27.1 / 40.8 & 20.5 / 38.7 & 31.0 / 45.3 & 28.9 / 48.5 & 41.6 / 59.5 & 28.6 / 44.8 & 21.3 / 45.2 & 12.7 / 28.2 \\
        ARC-Qwen-Video-Narrator$^*$ & 7B & 32.8 / 48.5 & 29.0 / 41.2 & 43.5 / 51.9 & 30.0 / 41.7 & 17.5 / 33.2 & 4.6 / 7.8 & 34.7 / 52.5 & 42.4 / 62.1 & 37.8 / 52.9 & 27.6 / 47.9 & 0.8 / 2.9 \\
        Qwen3-Omni-Instruct & 30B-A3B & 36.8 / 47.5 & 32.0 / 40.6 & 43.0 / 47.8 & 33.3 / 42.7 & 23.8 / 34.5 & 0.0 / 0.0 & 39.2 / 51.0 & 46.7 / 57.0 & 39.6 / 50.5 & 35.1 / 49.2 & 19.3 / 25.3 \\
        AVoCaDO & 7B & 38.7 / 51.7 & 33.9 / 45.4 & 48.7 / 54.4 & 30.6 / 41.3 & 27.0 / 42.6 & 28.6 / 51.1 & 41.4 / 55.3 & 39.4 / 52.1 & 43.5 / 53.1 & 41.8 / 60.7 & 31.9 / 38.3 \\
        Qwen3-Omni-Captioner & 30B-A3B & 43.9 / 58.8 & 41.0 / 53.5 & 56.1 / 60.6 & 38.5 / 51.7 & 32.5 / 49.7 & 43.6 / 63.1 & 45.4 / 61.7 & 49.3 / 62.2 & 49.6 / 61.3 & 40.5 / 62.8 & 29.6 / 44.7 \\
        \midrule
        DiaDem (Ours) & 7B & \textbf{65.9} / \textbf{79.3} & \textbf{57.2} / \textbf{70.3} & \textbf{74.2} / \textbf{79.1} & \textbf{60.6} / \textbf{72.4} & \underline{42.4} / \underline{62.2} & 55.7 / 68.3 & \textbf{70.9} / \textbf{84.4} & \textbf{76.9} / \textbf{87.5} & \underline{75.4} / \textbf{86.5} & 65.6 / \textbf{83.2} & 42.9 / 56.1 \\
    \bottomrule
    \end{tabular}
    }
    \caption{Model performance on DiaDemBench. $N$ denotes the speaker count. ``Overlap'' refers to subsets with temporally overlapping speech and is mutually exclusive with the groups defined by speaker count $N$. $^*$For ARC-Qwen-Video-Narrator, speaker and utterance information appears only in the thinking phase rather than the final answer, thus we use the thinking content as the model’s output for evaluation.}
    \label{tab:main result}
\end{table*}

\subsection{Data Annotation Pipeline for SFT}
Through our empirical case analysis, we observe complementary strengths between two Gemini variants: Gemini-2.5-Pro excels at transcribing utterances, whereas, when given comparable transcription performance, Gemini-3-Pro exhibits superior speaker attribution ability. Capitalizing on this observation, we design a data annotation pipeline, as illustrated in Fig.~\ref{fig:sft_grpo_data_curation}.~\label{sec:sft_pipeline}

First, Gemini-2.5-Pro is employed to generate raw dialogue descriptions with high utterance fidelity but relatively weak speaker attribution. These outputs are then refined by Gemini-3-Pro, which corrects speaker attribution, yielding 70K high-quality dialogue descriptions. In parallel, we utilize the holistic captioning capability of AVoCaDO to produce base audiovisual captions for these samples. Gemini-3-Pro then integrates the refined dialogue descriptions into these base captions, resulting in 70K high-quality audiovisual captions with more precise dialogue descriptions. In addition, we annotate 15K non-dialogue videos to preserve the model's general audiovisual captioning capability. Details regarding the video sources are provided in App.~\ref{app:video_source}.

We then perform SFT on AVoCaDO using the combined 85K dataset, equipping the model with foundational dialogue description skills while maintaining its general captioning performance.\label{sec:sft_anno}

\subsection{Difficulty-Partitioned Two-Stage GRPO}
Beyond SFT, we further enhance DiaDem through RL, specifically Group Relative Policy Optimization (GRPO), with 3K manually annotated high-quality dialogue descriptions. We utilize our proposed dialogue description metrics as the dialogue reward, combined with the checklist-based reward and length-regularized reward from the base model, to serve as the overall reward function for GRPO, which is detailed in App.~\ref{app:grpo}.\label{sec:grpo_anno}

Effective policy updates in GRPO rely on reward variations across multiple rollouts of the same sample. However, our preliminary experiments reveal that for samples that are either overly simple or excessively difficult, dialogue reward scores exhibit minimal variance across repeated inferences. This may lead to uninformative gradients that dilute the learning signal during policy optimization, thereby reducing training efficiency. To mitigate this issue, we first discard samples whose multiple rollouts yield nearly identical rewards due to simplicity, and adopt a difficulty-partitioned two-stage training strategy to progressively strengthen the model’s ability to learn from challenging instances.

Specifically, in Stage~1, the model is trained on the entire filtered dataset to improve generalization across samples of varying difficulty. Prior to Stage~2, we partition the filtered dataset based on the average dialogue rewards and isolate a subset of high-difficulty samples. In Stage~2, we double this high-difficulty subset to further strengthen the model’s performance on challenging cases. Additional details are provided in App.~\ref{app:two_stage_grpo}.

\section{Experiments}
\subsection{Experimental Settings}
We evaluate the dialogue description accuracy in audiovisual captioning of 14 representative models on DiaDemBench, including the Gemini series~\citep{comanici2025gemini}, video-SALMONN-2~\citep{tang2025video}, OmniVinci~\citep{ye2025omnivinci}, the ARC-Qwen-Video series~\citep{ge2025arc}, HumanOmniV2\citep{yang2025humanomniv2}, UGC-VideoCaptioner~\citep{wu2025ugc}, AVoCaDO~\citep{chen2025avocado}, the Qwen-Omni series~\citep{xu2025qwen2, Qwen3-Omni, ma2025omni}, as well as our proposed DiaDem.

In addition, we also assess the overall audiovisual captioning performance of DiaDem on the video-SALMONN-2 testset~\citep{tang2025video} and UGC-VideoCap~\citep{wu2025ugc}.

\subsection{Experimental Results}
\subsubsection{\hspace{-8pt}Evaluation \hspace{-1pt}of \hspace{-1pt}Dialogue \hspace{-1pt}Description \hspace{-1pt}Quality}
Tab.~\ref{tab:main result} presents the performance of various models on DiaDemBench in terms of their ability to accurately describe dialogues in audiovisual captions. The results demonstrate that the closed-source Gemini series substantially outperforms all previous open-source models in both speaker attribution and utterance transcription. However, a considerable gap with human performance remains. Among previous open-source models, the Qwen3-Omni series achieves leading results. AVoCaDO also attains competitive performance due to its holistic audiovisual captioning capabilities.

Notably, our proposed DiaDem, empowered by effective data construction and training strategies, surpasses the best-performing Gemini series by 2.3\% in overall speaker attribution and by 4.5\% in utterance transcription, and demonstrates consistent superiority in both single-shot and multi-shot scenarios. Nevertheless, under more challenging conditions such as scenes involving three or more speakers or overlapping speech, DiaDem still slightly lags behind Gemini, which constitutes one of our key directions for future improvement.

Further analysis and a detailed discussion of model behaviors across diverse dimensions of DiaDemBench are provided in App.~\ref{app:more_analysis}.

\begin{table}
    \centering
    \resizebox{\linewidth}{!}{
    \begin{tabular}{l@{\hspace{4pt}}c|c@{\hspace{5pt}}c@{\hspace{5pt}}c|c@{\hspace{4pt}}c@{\hspace{4pt}}c@{\hspace{4pt}}c}
    \toprule
        \multirow{2}{*}[-0.1cm]{\textbf{Model}} & \multirow{2}{*}[-0.1cm]{\textbf{Size}} & \multicolumn{3}{c|}{\textbf{SALMONN-2 testset}} & \multicolumn{4}{c}{\textbf{UGC-VideoCap}} \\
        \cmidrule(lr){3-5} \cmidrule(lr){6-9}
        ~ & ~ & Miss $\downarrow$ & Hall. $\downarrow$ & Total $\downarrow$ & Audio $\uparrow$ & Visual $\uparrow$ & Detail $\uparrow$ & Avg. $\uparrow$ \\
        \midrule
        \textcolor{gray!100}{Gemini-2.5-Pro} & \textcolor{gray!100}{-} & \textcolor{gray!100}{18.1} & \textcolor{gray!100}{13.3} & \textcolor{gray!100}{31.3} & \textcolor{gray!100}{69.5} & \textcolor{gray!100}{74.7} & \textcolor{gray!100}{73.7} & \textcolor{gray!100}{72.6} \\
         \textcolor{gray!100}{Gemini-3-Pro} &  \textcolor{gray!100}{-} &  \textcolor{gray!100}{21.4} &  \textcolor{gray!100}{14.2} &  \textcolor{gray!100}{35.6} &  \textcolor{gray!100}{69.7} &  \textcolor{gray!100}{75.6} &  \textcolor{gray!100}{74.3} &  \textcolor{gray!100}{73.2} \\
        \textcolor{gray!100}{Gemini-2.5-Flash} & \textcolor{gray!100}{-} & \textcolor{gray!100}{19.3} & \textcolor{gray!100}{13.9} & \textcolor{gray!100}{33.3} & \textcolor{gray!100}{69.1} & \textcolor{gray!100}{75.8} & \textcolor{gray!100}{74.0} & \textcolor{gray!100}{73.0} \\
        \midrule
        Qwen2.5-Omni & 7B & 41.7 & \underline{15.4} & 57.1 & 46.9 & 66.1 & 60.0 & 57.7 \\
        Qwen3-Omni-Instruct & 30B-A3B & 32.0 & \textbf{13.6} & 45.6 & 67.5 & 74.8 & \underline{72.3} & 71.5 \\
        Qwen3-Omni-Captioner & 30B-A3B & 31.0 & 16.6 & 47.6 & 69.0 & \underline{75.5} & \underline{72.3} & 72.5 \\
        UGC-VideoCaptioner & 3B & 31.6 & 17.0 & 48.6 & 61.4 & 58.4 & 57.5 & 59.1 \\
        video-SALMONN-2 & 7B & \underline{21.2} & 17.6 & 38.8 & 61.8 & 71.4 & 68.5 & 67.2 \\
        AVoCaDO & 7B & \textbf{21.1} & 16.2 & \underline{37.3} & \underline{73.0} & 74.6 & 71.8 & \underline{73.2} \\
        \midrule
        DiaDem (Ours) & 7B & \textbf{21.1} & 15.5 & \textbf{36.5} & \textbf{75.4} & \textbf{76.8} & \textbf{74.6} & \textbf{75.6} \\
    \bottomrule
    \end{tabular}
    }
    \caption{Model performance on the audiovisual video captioning benchmarks. Following AVoCaDO, we replace the judge model for the SALMONN-2 testset with GPT-4.1 to ensure more reliable evaluation.}
    \label{tab:caption result}
\end{table}

\subsubsection{\hspace{-4pt}Evaluation of Overall Captioning Quality}
To verify whether DiaDem improves dialogue description capability without compromising other aspects of audiovisual captioning, we evaluate its overall performance on two benchmarks specifically designed to measure holistic audiovisual captioning quality: the video-SALMONN-2 testset and UGC-VideoCap, as summarized in Tab.~\ref{tab:caption result}.

The results show that, while substantially enhancing dialogue description accuracy, DiaDem does not degrade other aspects of audiovisual captioning. On the contrary, it further improves the captioning quality of the base model and even outperforms the strongest Gemini-series model on UGC-VideoCap by 2.4\%.

\subsection{Ablation Studies}
\subsubsection{Ablation on the Post-Training Pipeline}
Tab.~\ref{tab:abla pipeline} presents a comprehensive ablation study of each component within our post-training pipeline.

Benefiting from our high-quality dialogue-aware audiovisual captioning data construction strategy, the SFT stage substantially enhances the base model’s ability to describe dialogues and also improves overall audiovisual caption quality on UGC-VideoCap. However, on the video-SALMONN-2 testset, performance degrades to a level comparable to the SFT version of the base model without GRPO, which scores 41.4 in the original paper. We hypothesize that this degradation arises from a mismatch in training paradigms, where SFT may override the performance gains established during the earlier GRPO stage. Subsequent application of our difficulty-partitioned two-stage GRPO effectively mitigates this issue, yielding steady improvements in both dialogue description accuracy and general audiovisual captioning quality.

\begin{figure*}[t]
  \includegraphics[width=\linewidth]{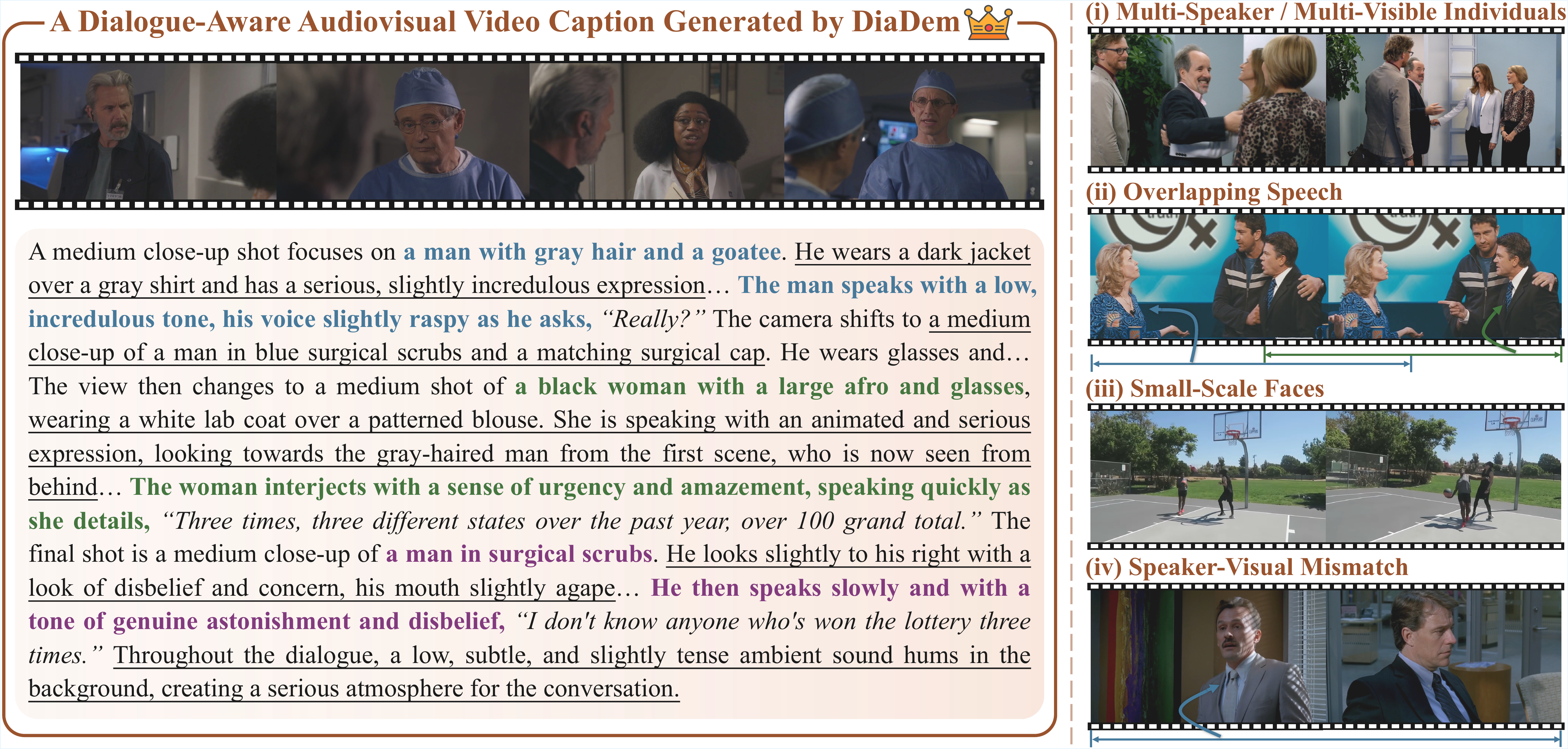}
  \caption{On the left, we present an illustration of an audiovisual video caption with accurate dialogue descriptions generated by DiaDem, featuring both \textbf{correct speaker attribution} and \textit{precise utterance transcription}, as well as other \underline{general audiovisual details}. On the right, we showcase four representative dialogue scenarios from DiaDemBench that are commonly challenging for existing models to produce accurate dialogue descriptions, with the aim of providing insights for future advancements in audiovisual video captioning models.}
  \label{fig:main_case}
\end{figure*}

To further dissect the design choices within the GRPO stage, we conduct additional ablations: (i) ``w/o staged training'' merges data from both stages into a single training phase; (ii) ``w/o easy filtering'' disables the filtering mechanism that discards overly simple samples; and (iii) ``w/o dialogue reward'' removes our proposed dialogue reward. All three variants underperform the difficulty-partitioned two-stage GRPO to varying degrees, validating the effectiveness of our strategy.

Additionally, to investigate whether the performance gains from GRPO are merely attributed to the increased data volume, we adapt the SFT data construction pipeline outlined in Sec.~\ref{sec:sft_pipeline} to convert the human-annotated dialogue descriptions in the GRPO dataset into dialogue-aware audiovisual captions for additional SFT. However, this modification even leads to a slight performance drop relative to the original SFT model, indicating that the gains from GRPO primarily stem from the reward design and staged training strategy, rather than from exposure to more data.

\begin{table}
    \centering
    \resizebox{\linewidth}{!}{
    \begin{tabular}{l|c@{\hspace{2pt}}c@{\hspace{-2pt}}c}
    \toprule
        \textbf{Model} & \textbf{\makecell{DiaDem-\\Bench}} & \textbf{\makecell{SALMONN-2\\testset $\downarrow$}} & \textbf{\makecell{UGC-\\VideoCap}} \\
        \midrule
        AVoCaDO (base model) & 38.7 / 51.7 & 37.3 & 73.2 \\
        \midrule
        \textcolor{gray!100}{\textit{Staged Post-training Recipe}} \\
        \quad + SFT & 59.3 / 74.0 & 42.0 & 74.8 \\
        \quad\quad + GRPO-Stage-1 & 64.7 / 77.6 & 38.4 & 74.7 \\
        \quad\quad\quad + GRPO-Stage-2 (DiaDem) & \textbf{65.9} / \textbf{79.3} & \textbf{36.5} & \textbf{75.6} \\
        \midrule
        \textcolor{gray!100}{\textit{GRPO Variants}} \\
        \quad\quad + GRPO w/o staged training & 64.5 / 78.2 & 37.5 & 74.8 \\
        \quad\quad + GRPO w/o easy filtering & 63.2 / 78.2 & 37.7 & 74.5 \\
        \quad\quad + GRPO w/o dialogue reward & 60.1 / 74.3 & 37.0 & 75.2 \\
        \midrule
        \textcolor{gray!100}{\textit{w/o GRPO}} \\
        \quad\quad + SFT using GRPO data & 59.2 / 73.6 & 42.4 & 75.0 \\
    \bottomrule
    \end{tabular}
    }
    \caption{Ablation study on our post-training pipeline.}
    \label{tab:abla pipeline}
\end{table}

\begin{table}
    \centering
    \resizebox{\linewidth}{!}{
    \begin{tabular}{l|c@{\hspace{2pt}}c@{\hspace{0pt}}c}
    \toprule
        \textbf{Model} & \textbf{\makecell{DiaDem-\\Bench}} & \textbf{\makecell{SALMONN-2\\testset $\downarrow$}} & \textbf{\makecell{UGC-\\VideoCap}} \\
        \midrule
        \multicolumn{4}{l}{\textcolor{gray!100}{\textit{GRPO w/ the problematic dialogue reward in AVoCaDO}}} \\
        \quad Stage-1 & 63.5 / 76.9 & 40.1 & 74.6 \\
        \quad Stage-2 & 63.7 / 77.2 & 37.1 & 74.9 \\
        \midrule
        \multicolumn{4}{l}{\textcolor{gray!100}{\textit{GRPO w/ our enhanced dialogue reward}}} \\
        \quad Stage-1 & 64.7 / 77.6 & 38.4 & 74.7 \\
        \quad Stage-2 (DiaDem) & \textbf{65.9} / \textbf{79.3} & \textbf{36.5} & \textbf{75.6} \\
    \bottomrule
    \end{tabular}
    }
    \caption{Ablation on our enhanced dialogue reward.}
    \label{tab:abla origianl reward}
\end{table}

\subsubsection{\hspace{-8.8pt}Ablation \hspace{-1.18pt}on \hspace{-1.18pt}the \hspace{-1.18pt}Enhanced \hspace{-1.18pt}Dialogue \hspace{-1.18pt}Reward}
In Sec.~\ref{sec: Evaluation Protocol}, we identify key limitations of the dialogue reward used in AVoCaDO and propose targeted improvements. As shown in Tab.~\ref{tab:abla origianl reward}, our enhanced dialogue reward consistently outperforms the original formulation across both stages of GRPO training, yielding not only more accurate dialogue descriptions but also enhanced overall audiovisual captioning performance.

\subsection{Qualitative Analysis}
In the left panel of Fig.~\ref{fig:main_case}, we present an illustrative example of a dialogue-aware audiovisual video caption generated by DiaDem. The result demonstrates DiaDem's capability to produce accurate dialogue descriptions with both correct speaker attribution and precise utterance transcription, while simultaneously delivering strong descriptive performance regarding other audiovisual details within the scene. Additional qualitative comparisons between DiaDem and two strong Gemini series models are provided in App.~\ref{app:additional_cases}.

In the right panel of Fig.~\ref{fig:main_case}, we highlight four representative dialogue scenarios from DiaDemBench that remain challenging for state-of-the-art audiovisual video captioning models. Beyond the commonly acknowledged difficulties such as scenes with multiple speakers, multiple visible individuals, or overlapping speech, our extensive case studies identify several less-explored yet critical failure cases that frequently lead to erroneous descriptions in current models. These include situations where speakers occupy only small facial regions, as well as cases involving mismatches between the off-screen speaker and the characters currently visible on the screen. We hope that these observations can provide valuable insights for the future development of more robust and reliable audiovisual video captioning models.

\section{Conclusion}
This paper focuses on the critical yet underexplored challenge of accurately describing dialogues in audiovisual video captioning. We first design a suite of evaluation metrics to faithfully capture two core components of dialogue descriptions: speaker attribution and utterance transcription. Based on these metrics, we construct DiaDemBench, which features diverse dialogue scenarios and enables a comprehensive and reliable evaluation of dialogue description fidelity in audiovisual captions.

Evaluations on DiaDemBench reveal a substantial performance gap between existing open-source and commercial models. To bridge this gap, we propose DiaDem, an audiovisual video captioning model capable of both utterance transcription and speaker attribution, while preserving holistic audiovisual context. Building upon AVoCaDO, we first synthesize a high-quality dataset for SFT and then incorporate human-annotated samples to perform a difficulty-partitioned two-stage GRPO to further enhance dialogue description quality. Experimental results show that DiaDem not only outperforms the Gemini series in dialogue description accuracy, but also achieves competitive performance on general audiovisual captioning benchmarks. Comprehensive ablation studies further confirm the contribution of each component in our training pipeline, underscoring the effectiveness of our approach.

\section*{Limitations}
Although DiaDem achieves state-of-the-art performance in generating dialogue-aware audiovisual video captions, it still falls short of human-level performance, particularly in complex scenarios involving multi-party interactions and overlapping speech. Addressing these challenges represents a critical avenue for future research.

\section*{Ethical Considerations}
While DiaDem improves the accuracy of dialogue descriptions in audiovisual video captioning, it is not infallible. The model may occasionally produce incorrect speaker attributions or hallucinated utterance, particularly in challenging cases of overlapping speech or multi-party interactions. Such inaccuracies could lead to serious misunderstandings, defamation, or the spread of misinformation. Therefore, DiaDem should not be deployed as the sole authoritative source in high-stakes applications, such as legal transcription or evidence analysis, without thorough human verification.


\bibliography{custom}

\clearpage
\appendix

\twocolumn[{%
  \centering
  \includegraphics[width=0.8\linewidth]{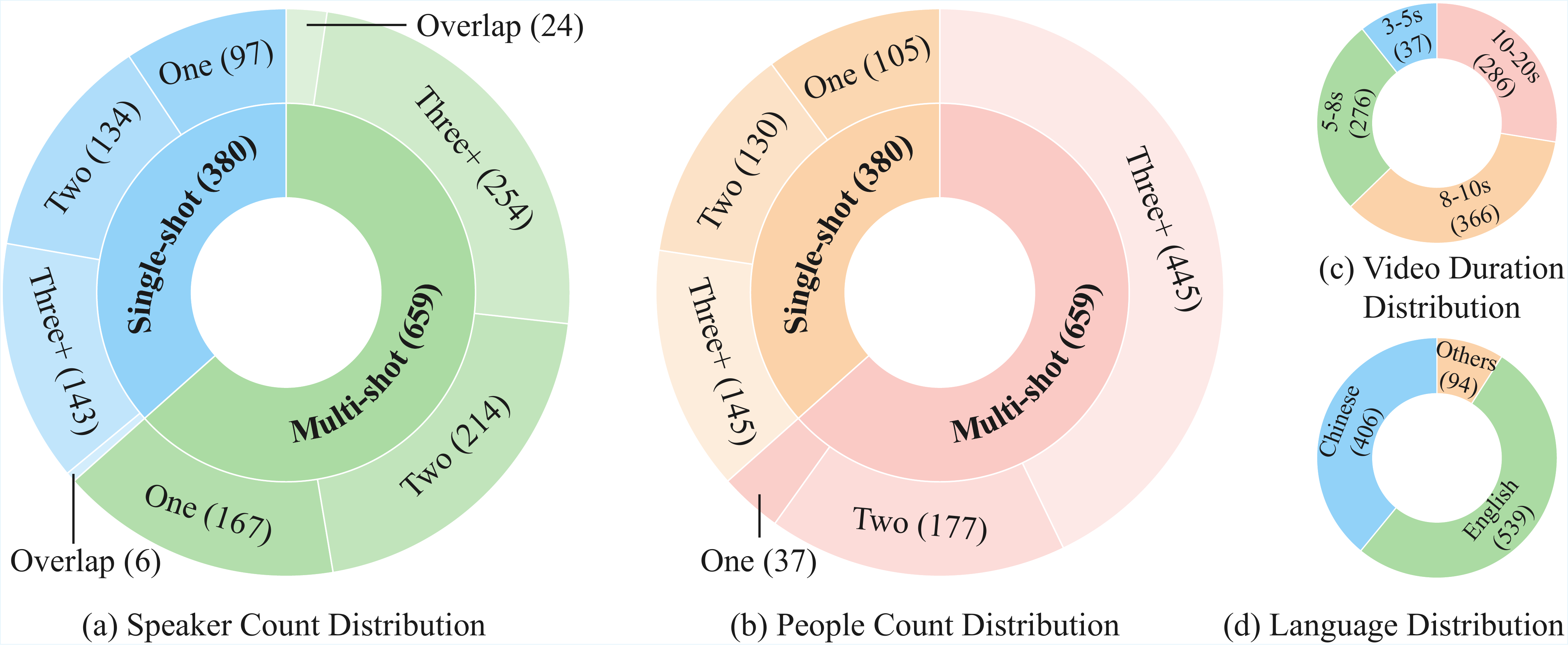}
  \captionof{figure}{Dataset Statistics for DiaDemBench.}
  \label{fig:statistics}
  \vspace{1em}
  \includegraphics[width=\linewidth]{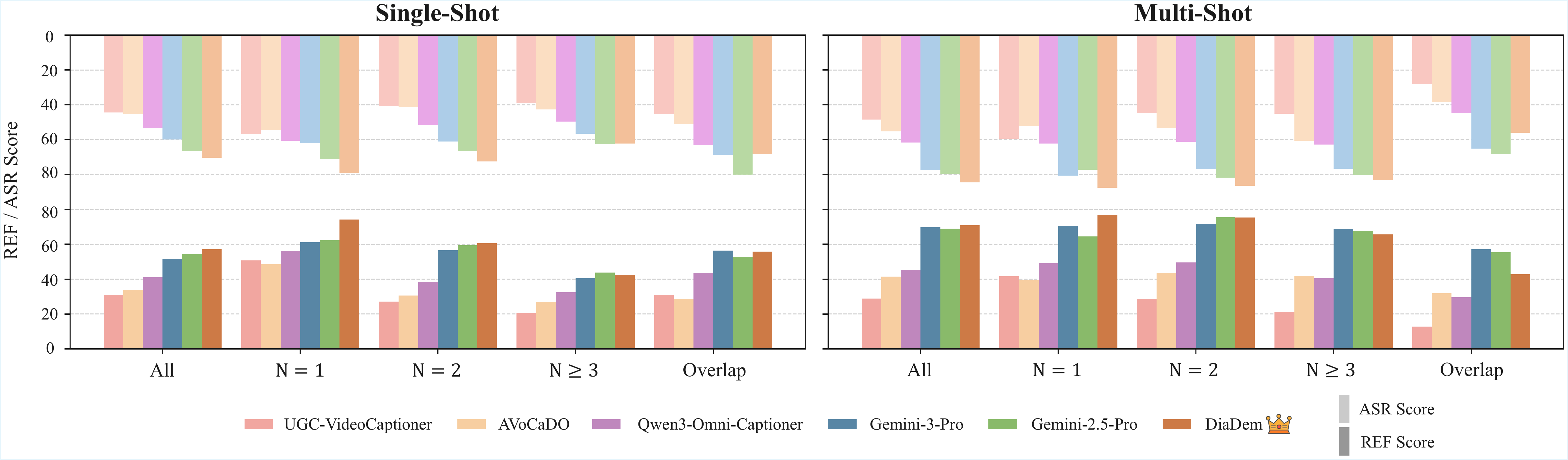}
  \captionof{figure}{Analysis of model performance across varying speaker counts $N$. ``Overlap'' refers to subsets with temporally overlapping speech and is mutually exclusive with the groups defined by speaker count.}
  \label{fig:analy_speaker_num}
  \vspace{2em}
}]

\section{Details of DiaDemBench}
\subsection{Dataset Statistics}
Fig.~\ref{fig:statistics} provides a holistic statistical overview of DiaDemBench across several key dimensions.

First, in terms of shot editing types, DiaDemBench comprises 37\% single-shot and 63\% multi-shot videos, with the latter exhibiting more scene transitions. Within each shot category, the number of speakers is relatively balanced, with multi-speaker scenarios accounting for the majority (Fig.~\ref{fig:statistics}\textcolor[HTML]{000080}{a}). To further challenge models' dialogue description capabilities, we also incorporate a small proportion of videos featuring overlapping speech.

To modulate the difficulty of speaker attribution, we regulate the distribution of on-screen characters such that videos containing multiple visible individuals constitute the majority (Fig.~\ref{fig:statistics}\textcolor[HTML]{000080}{b}). It should be noted that in clips featuring only one visible character, off-screen speakers may still be present; thus, speaker attribution is not necessarily straightforward even in such seemingly simple cases.

Fig.~\ref{fig:statistics}\textcolor[HTML]{000080}{c} illustrates the distribution of video durations. To enable most models to capture speaker-related details (e.g., lip movements or gestures) at relatively high resolution and frame rate, we restrict all clips to under 20 seconds, with durations following an approximately uniform distribution.

Finally, Fig.~\ref{fig:statistics}\textcolor[HTML]{000080}{d} displays the language distribution in DiaDemBench. English and Chinese account for the majority, while Japanese, Korean, French, and several other languages are also included to ensure broad linguistic diversity.

\begin{figure*}[t]
  \includegraphics[width=\linewidth]{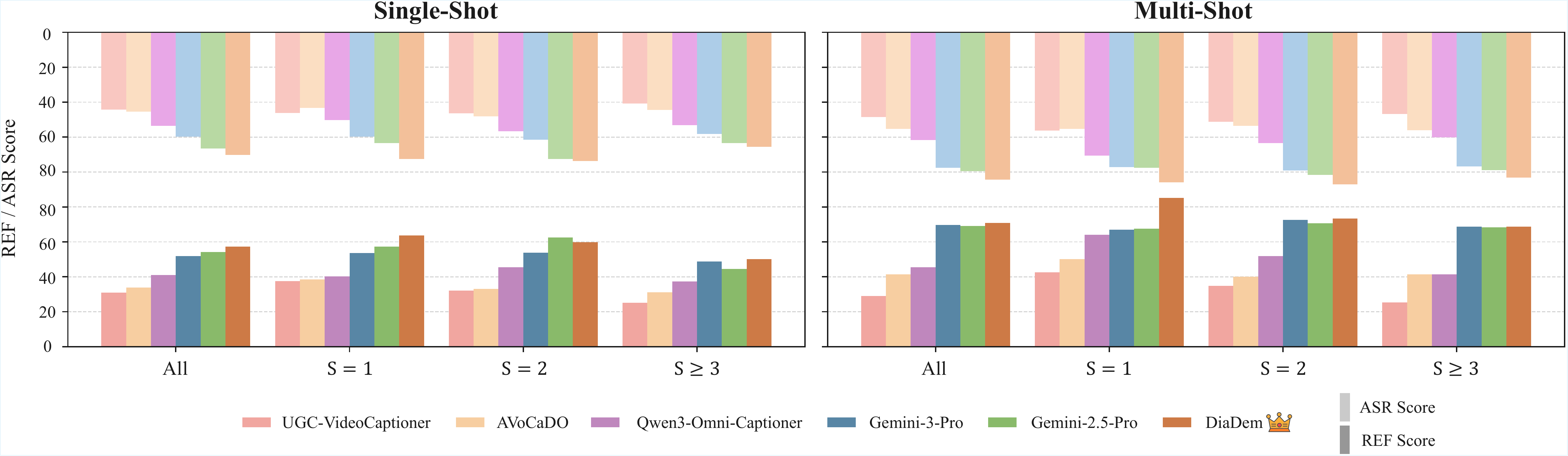}
  \caption{Analysis of model performance across varying numbers of on-screen individuals $S$.}
  \label{fig:analy_people_num}
\end{figure*}

\begin{figure*}[t]
  \includegraphics[width=\linewidth]{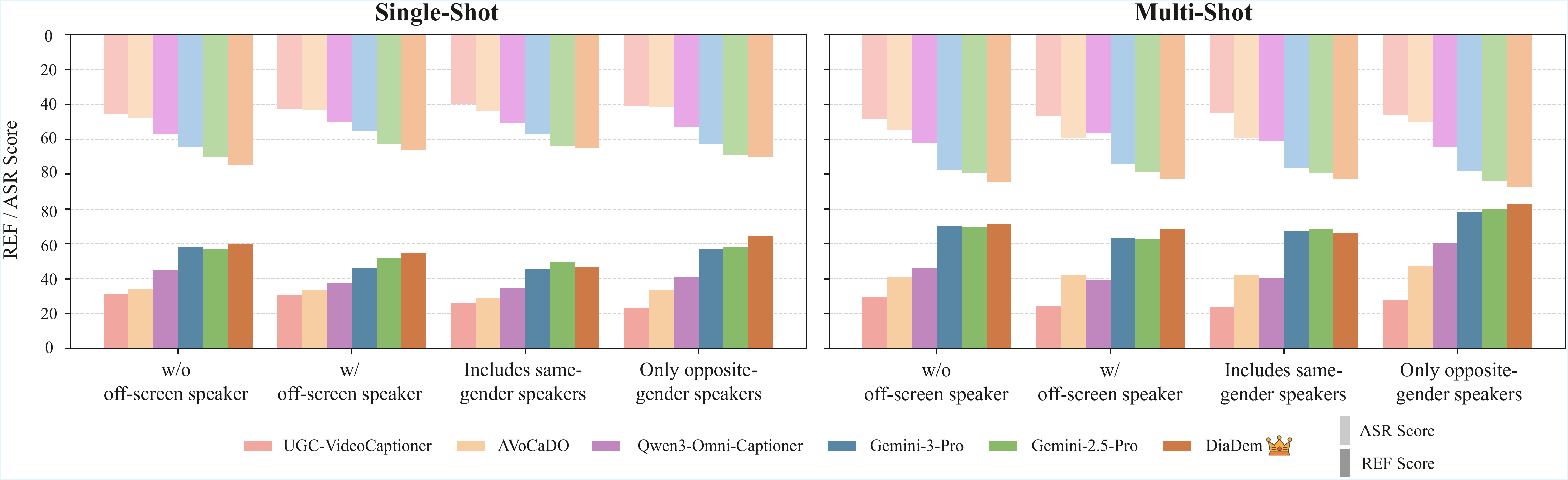}
  \caption{Analysis of model performance regarding the presence of off-screen speaker and speaker gender.}
  \label{fig:analy_others}
\end{figure*}

\subsection{Further Analysis}
In this subsection, we present a multi-dimensional analysis of several representative models on DiaDemBench, aiming to identify key challenges and offer potential directions for future research in dialogue-aware audiovisual video captioning.\label{app:more_analysis}

\subsubsection{Model Performance Across Varying Speaker Counts}
In Fig.~\ref{fig:analy_speaker_num}, we compare model performance under single-shot and multi-shot settings across scenarios with varying numbers of speakers. Our observations reveal the following insights:

(i) Except for our proposed DiaDem, all open-source models consistently underperform the closed-source Gemini series across all dimensions.

(ii) Models generally achieve higher performance in multi-shot scenarios than in single-shot ones. Through qualitative case analysis, we observe that this advantage may be attributed to the fact that multi-shot videos often featuring close-ups of the speakers in each shot, which clarify speaker references and facilitates identification. Moreover, the visual variations across shots provide implicit alignment anchors for ASR, making it easier to segment speech for each speaker and thus improving transcription accuracy.

(iii) As the number of speakers increases, model performance degrades in both speaker attribution and utterance transcription. Notably, DiaDem also underperforms the Gemini series in scenarios with three or more speakers, highlighting a key direction for future work: improving dialogue description capabilities in complex multi-speaker settings.

(iv) In videos with overlapping speech, both open-source and commercial models generally perform poorly, indicating a need for stronger audio source separation capabilities, as well as tighter integration of vocal timbre and visual cues to accurately attribute utterances to corresponding speakers, representing another key area for future work.

\subsubsection{Model Performance Across Varying Numbers of On-Screen Individuals}
Fig.~\ref{fig:analy_people_num} presents performance comparisons across scenarios with varying numbers of on-screen individuals. Two salient patterns emerge from them:

(i) In single-shot scenarios, dialogue captioning performance consistently declines as the number of visible individuals increases.

(ii) In multi-shot settings, however, performance remains relatively stable between scenes containing two versus three or more on-screen individuals.

This discrepancy may be attributed to the cinematographic conventions. In multi-shot sequences, the prevalent use of speaker-focused close-ups isolates the active speaker, thereby mitigating the visual complexity introduced by additional on-screen individuals. In contrast, single-shot videos with multiple people typically feature smaller facial regions and visually crowded scenes, making accurate speaker identification and utterance transcription substantially more challenging.

\subsubsection{Model Performance Across Other Dimensions}
Additional observations are presented in Fig.~\ref{fig:analy_others}:

(i) Models exhibit weaker performance on videos containing off-screen speakers. The absence of corresponding visual cues often leads models to misattribute voice-over dialogue to visible characters, resulting in erroneous dialogue descriptions.

(ii) Performance also drops significantly in videos containing same-gender speakers compared to those with only mixed-gender ones, where natural differences in vocal timbre provide useful discriminative signals.

These results indicate that current models remain limited in their ability to jointly exploit audio and visual modalities for accurate speaker-utterance alignment. Future work could prioritize more effective fusion strategies that leverage both vocal characteristics and contextual visual information to resolve speaker identity in diverse and challenging audiovisual scenarios.

\subsection{Ablation on the Judge Model}
In the main experiments, we use Gemini-2.5-Pro to extract structured dialogue descriptions from audiovisual captions, and then employ Gemini-2.5-Flash to assess the consistency of speaker descriptions within successfully matched dialogue tuples, thereby obtaining the speaker reference accuracy score. To account for scenarios where closed-source model APIs are unavailable, and to further assess the generalizability of our evaluation protocol with respect to the choice of judge model, we conduct an ablation study by replacing the Gemini series with Qwen3-VL-32B~\citep{Qwen3-VL} for evaluation. The results are reported in Tab.~\ref{tab:abla judge}.

The experimental results reveal that, although the absolute scores produced by different judge models exhibit minor fluctuations, which may stem from inherent model-specific biases, the relative ranking among the evaluated models remains largely consistent. This indicates that our evaluation protocol is not overly sensitive to the specific choice of judge model. As long as the judge model possesses strong capabilities and can deliver stable, fair judgments, it is suitable for integration into DiaDemBench. Additionally, aggregating judgments from multiple diverse judge models can help mitigate individual model biases, thereby yielding a more reliable and robust evaluation, albeit at an increased computational cost.

\begin{table}
    \centering
    \resizebox{\linewidth}{!}{
    \begin{tabular}{l|cc}
    \toprule
        \multirow{2}{*}{\textbf{Model}} & \textbf{Gemini Series$^*$} & \textbf{Qwen3-VL-32B} \\
        ~ & REF / ASR & REF / ASR \\
        \midrule
        Gemini-2.5-Pro & \underline{63.6} / \underline{74.8} & 59.7 / \underline{71.5} \\
        Gemini-3-Pro & 63.1 / 71.0 & \underline{61.3} / 69.1 \\
        Gemini-2.5-Flash & 58.5 / 73.3 & 58.0 / \underline{71.5} \\
        \midrule
        video-SALMONN-2 & 11.5 / 16.6 & 11.0 / 15.5 \\
        Qwen2.5-Omni & 26.1 / 37.1 & 24.8 / 34.6 \\
        UGC-VideoCaptioner & 29.7 / 47.0 & 30.5 / 45.0 \\
        ARC-Qwen-Video-Narrator & 32.8 / 48.5 & 27.6 / 35.3 \\
        Qwen3-Omni-Instruct & 36.8 / 47.5 & 36.7 / 46.2 \\
        AVoCaDO & 38.7 / 51.7 & 36.5 / 50.3 \\
        Qwen3-Omni-Captioner & 43.9 / 58.8 & 43.8 / 58.7 \\
        \midrule
        DiaDem (Ours) & \textbf{65.9} / \textbf{79.3} & \textbf{63.4} / \textbf{78.3} \\
    \bottomrule
    \end{tabular}
    }
    \caption{Ablation study on the judge model. $^*$In main experiments, we use Gemini-2.5-Pro for dialogue extraction and Gemini-2.5-Flash for speaker accuracy evaluation to balance cost and accuracy.}
    \label{tab:abla judge}
\end{table}

\subsection{Data Curation}\label{app:anno_pipeline}
\subsubsection{Video Collection}
Open-source video datasets are often carefully curated and highly valuable for in-depth analysis. However, they are frequently extensively captioned and may have been incorporated into the training corpora of existing models. To minimize data leakage and ensure that most of videos remain unseen during prior training, we collect data from publicly available UGC platforms.

Specifically, to evaluate the accuracy of dialogue descriptions in audiovisual video captioning under more diverse scenarios, we select movie clips from trailers and short-form videos that contain rich dialogue scenes, as identified through metadata filtering. To respect copyright constraints, our benchmark will be released under highly restrictive licensing terms, permitting its use exclusively for academic research purposes.

Considering the context window limitations of current models, raw videos are segmented into clips of no more than 20 seconds using PySceneDetect. Subsequently, multiple open-source models are employed to infer and cross-validate key attributes of each clip, including speaker counts, the number of visible individuals, language information, and shot editing types. Based on these attributes, we further filter the videos to ensure broad category coverage and balanced attribute distributions.

Finally, each video clip is manually reviewed to exclude offensive content and to ensure suitability for academic research. Through this process, we obtain a total of 1,039 video segments for subsequent annotation.

\begin{figure*}[t]
  \includegraphics[width=\linewidth]{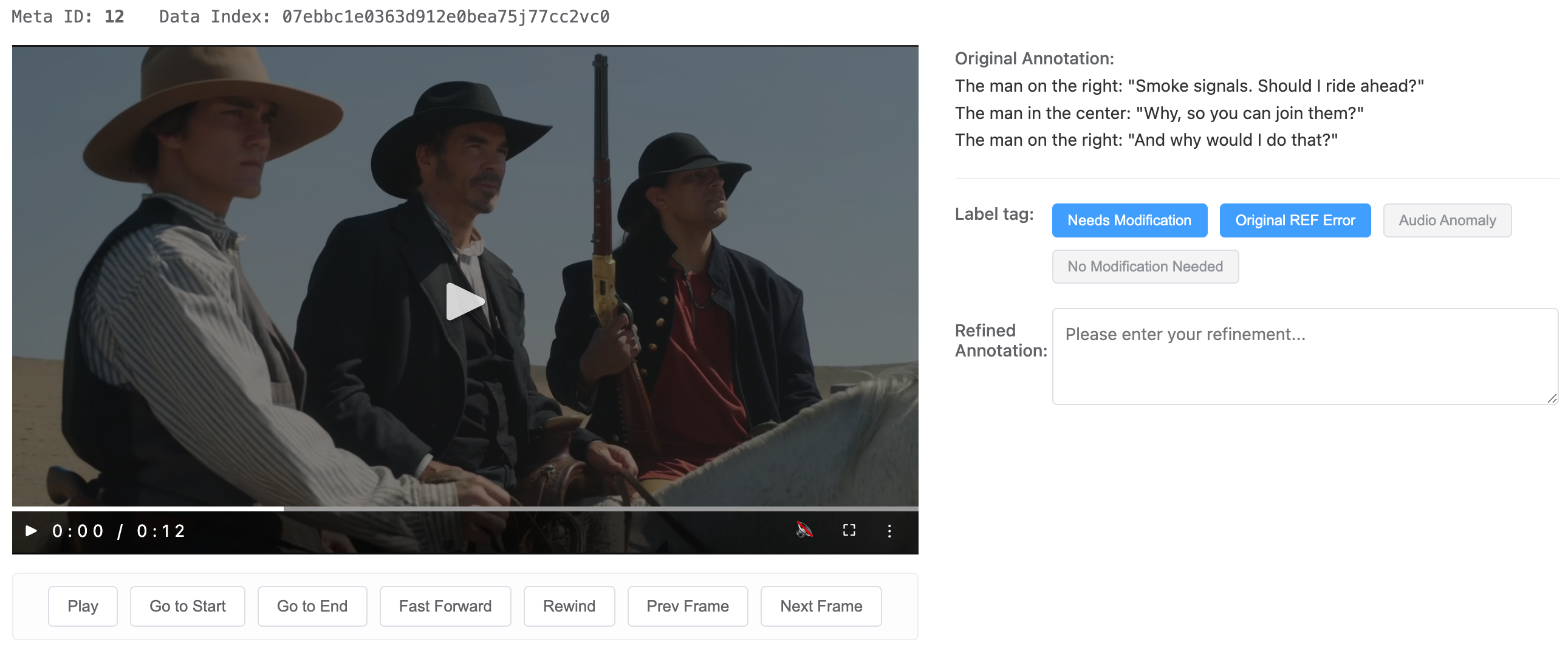}
  \caption{Screenshot of the annotation system interface.}
  \label{fig:label_interface}
\end{figure*}

\subsubsection{Annotation Pipeline}
When constructing DiaDemBench, in order to obtain accurate ground-truth dialogue annotations efficiently, we adopt a hybrid annotation pipeline that combines automatic annotation with subsequent manual refinement. Specifically, we first leverage Gemini-2.5-Pro to produce initial dialogue descriptions for each video clip. However, these machine-generated annotations frequently suffer from speaker attribution errors and utterance transcription inaccuracies. To ensure annotation quality, a team of professionally trained annotators is employed to meticulously revise the initial annotations. A sample is accepted only when all three annotators reach consensus, as verified by a senior supervisor. Any remaining disagreements are adjudicated by the senior supervisor, thereby ensuring reliable annotation quality.

\subsection{Human Annotators}
We recruit ten experienced multilingual annotators based in Asia through a crowdsourcing platform to participate in the annotation process. To illustrate the annotation workflow, we provide a screenshot of the annotation interface in Fig.~\ref{fig:label_interface}, which demonstrates how annotators interact with the system and perform labeling tasks. Specifically, annotators are required to watch the original videos and, referring to the initial machine-generated dialogue descriptions, assign appropriate labels and provide refined dialogue descriptions.

To ensure the quality and reliability of the annotations, annotators are compensated based on the time spent rather than the number of samples completed, thereby minimizing incentives for rushed or superficial work. Annotators are paid at a rate of USD~10 per hour, which is highly competitive relative to prevailing industry standards for comparable annotation tasks.

\subsection{Additional Implementation Details}
In this subsection, we provide further implementation details concerning the matching of dialogue lists. Specifically, we first apply regular expressions to normalize the text by removing punctuation and whitespace, and all Latin characters are converted to lowercase. For Traditional Chinese text, we employ the OpenCC\footnote{\url{https://github.com/BYVoid/OpenCC}} to convert it into Simplified Chinese. These preprocessing steps ensure that the matching process focuses on the accuracy of lexical accuracy, rather than being affected by superficial formatting variations. The similarity threshold $\gamma$ for utterance matching is set to 0.6, and the maximum merging window size $W$ is set to 6.
\label{app:asr_detail}

\section{Details of DiaDem}
\subsection{Video Sources}\label{app:video_source}
In this subsection, we describe the video sources used to train DiaDem. The training data consist of 70K dialogue-rich videos and 15K non-dialogue videos for SFT, along with an additional 3K dialogue-rich videos for GRPO.

To strengthen the model’s ability to generate audiovisual video captions with precise dialogue descriptions, we curate a diverse collection of videos featuring rich verbal interactions from publicly available film clips and UGC datasets. Specifically, during the SFT stage, we sample 8K videos from CinePile~\citep{rawal2024cinepile}, 21K from YouTube-Commons~\citep{Youtube-Commons}, 26K from OMEGA~\citep{OMEGA}, 10K from TikTok-10M~\citep{tiktok_10m_2025}, and 5K from Short-Video~\citep{shang2025large}. In the subsequent GRPO stage, we use an additional 1.8K CinePile videos that are distinct from those used in the SFT, together with 0.7K videos from Social-IQ~\citep{siq2} and 0.5K from Friends-MMC~\citep{wang2025friends}. All above videos are pre-filtered by specialized expert models to verify the presence of human speech, ensuring that the model is effectively trained to describe dialogues when generating audiovisual captions.

Meanwhile, to preserve the model’s captioning capability in non-dialogue scenarios, we additionally incorporate 11K non-dialogue videos from VGGSound~\citep{chen2020vggsound} and 4K from Music-AVQA~\citep{li2022learning} during SFT. All of the above datasets are licensed for academic research use. Detailed annotation procedures for different video categories are described in Secs.~\ref{sec:sft_anno} and~\ref{sec:grpo_anno}.

\subsection{Group Relative Policy Optimization\label{app:grpo}}

\subsubsection{Formulation}
Group Relative Policy Optimization (GRPO) substantially improves computational efficiency by eliminating the need for a separate critic model in Proximal Policy Optimization (PPO). Instead of estimating absolute value functions, GRPO leverages relative rewards within a group of sampled responses to compute advantages. Specifically, for each input query $q$, GRPO works draws a group of $G$ responses $\{ o_1, o_2, ..., o_G \}$ from the old policy model $\pi_{\theta_{old}}$, then computing their corresponding rewards $\{ r_1, r_2, ..., r_G \}$ to derive the advantage function $A_i$ for response $o_i$:
$$A_i = \frac{r_i - \text{mean}(\{r_1, r_2, \dots, r_G\})}{\text{std}(\{r_1, r_2, \dots, r_G\})}$$
The current policy model $\pi_{\theta}$ is then updated by maximizing the following objective function:
$$
    \hspace*{-18pt}
    \begin{aligned}
    \mathcal{J}&_{\text{GRPO}}(\theta) = \mathbb{E}_{\{o_i\}_{i=1}^G \sim \pi_{\theta_{\text{old}}}(o_i|q)} \Bigg[ \frac{1}{G} \sum_{i=1}^G \bigg( \min \Big( r_i(\theta) A_i, \\
    &\text{clip} \Big( r_i(\theta), 1-\varepsilon, 1+\varepsilon \Big) A_i \Big) - \beta \cdot \mathbb{D}_{\text{KL}} \left( \pi_\theta || \pi_{\text{ref}} \right) \bigg) \Bigg]
    \end{aligned}
    \label{eq:grpo}
$$
where $r_i(\theta)=\frac{\pi_\theta(o_i|q)}{\pi_{\theta_{\text{old}}}(o_i|q)}$ denotes the importance sampling ratio, $\epsilon$ is the clipping parameter that restricts policy updates within a trust region, $\beta$ adjusts the strength of the KL divergence penalty, and $\pi_{\text{ref}}$ is the reference policy model employed to enhance training stability.

\begin{figure*}[t]
  \includegraphics[width=\linewidth]{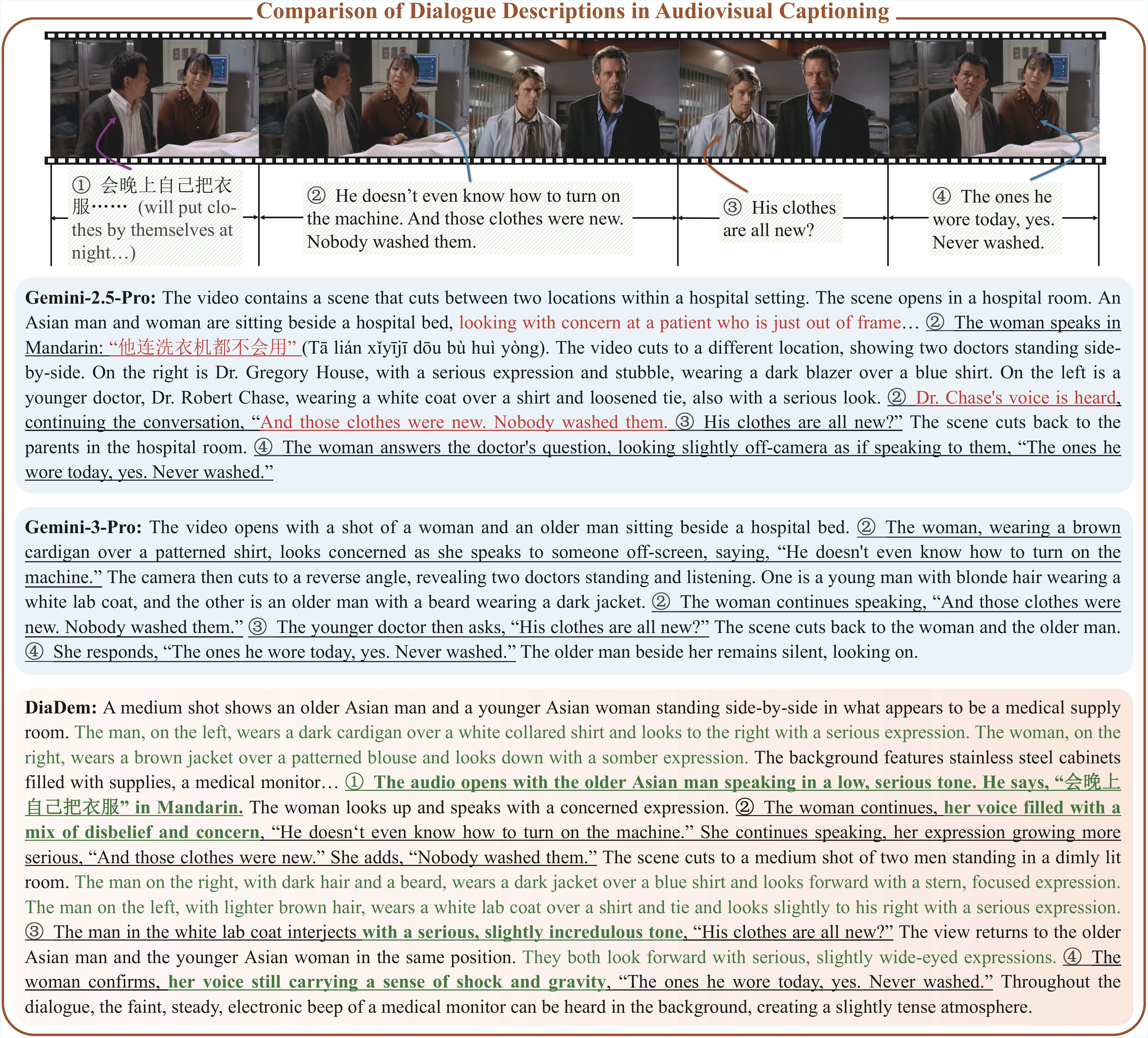}
  \caption{Qualitative comparison of DiaDem against two strong Gemini series models: Gemini-2.5-Pro and Gemini-3-Pro. Dialogue descriptions in the audiovisual captions are \underline{underlined}, and circled indices indicate correspondence with the respective ground-truth dialogues, aiding in the identification of omissions.}
  \label{fig:app_case_1}
\end{figure*}

\subsubsection{Reward Functions}
In the difficulty-partitioned two-stage GRPO strategy, we employ three complementary reward functions. The first is the dialogue reward $\mathcal{R_\mathrm{D}}$, defined as the average of the advanced dialogue description evaluation metrics proposed in Sec.~\ref{sec: Evaluation Protocol}:
$$\mathcal{R_\mathrm{D}}=(\mathrm{REF}+\mathrm{ASR}) \mathbin{\big/}  2$$
In addition, we incorporate the checklist-based reward $\mathcal{R_\mathrm{C}}$ and the length-regularized reward $\mathcal{R_\mathrm{L}}$ from AVoCaDO, which encourage the completeness of audiovisual descriptions and regulate caption length, respectively. The final reward $\mathcal{R}$ used during training is defined as the sum of these three components:
$$\mathcal{R}=\mathcal{R_\mathrm{D}}+\mathcal{R_\mathrm{C}}+\mathcal{R_\mathrm{L}}$$

\subsubsection{\hspace{-6pt}Difficulty-Partitioned\hspace{-0.5pt} Two-Stage\hspace{-0.5pt} Strategy}\label{app:two_stage_grpo}
This subsection details the difficulty-partitioned two-stage GRPO strategy introduced in Sec.~\ref{sec:grpo_anno}.

To prevent effective learning signals from being diluted by uninformative gradients, we pre-filter overly simple samples whose dialogue rewards exhibit negligible variance across multiple rollouts prior to Stage~1. Concretely, starting from the original 3K manually annotated dataset, we generate eight independent rollouts per sample using the model after SFT and compute the mean and standard deviation of the dialogue reward $\mathcal{R}_{\mathrm{D}}$. Samples satisfying both $\text{mean}(\mathcal{R}_{\mathrm{D}}) > 0.8$ and $\text{std}(\mathcal{R}_{\mathrm{D}}) < 0.1$ are discarded as overly easy, resulting in 2.1K samples used for Stage~1 training.

Before Stage~2, we apply the model trained in Stage~1 to generate eight additional rollouts for each of the retained 2.1K samples. Samples with $\text{mean}(\mathcal{R}_{\mathrm{D}}) < 0.3$ are identified as challenging cases, resulting in a high-difficulty subset of 0.4K samples. During Stage~2, this high-difficulty subset is duplicated to further enhance the model’s performance on challenging dialogue scenarios.

\subsection{Implementation Details}
During the SFT stage, the model is trained for 2 epochs with a batch size of 128 and a learning rate of $2 \times 10^{-5}$. In the difficulty-partitioned two-stage GRPO phase, we use a batch size of 64 and a learning rate of $1 \times 10^{-5}$. For each input query, we sample 8 responses using a temperature of 1.0. The KL-divergence regularization coefficient $\beta$ is set to 0.02. Throughout training, both the video and audio encoders are kept frozen, and only the adapters and the LLM backbone are updated.

During both training and evaluation, video inputs are sampled at 2 fps, and the resolution of each frame is limited to a maximum of $512 \times 28 \times 28$ pixels. Due to the base model’s context window limitation of 32K tokens, the total video tokens is restricted to $25600 \times 28 \times 28$. All training is conducted on 16 NVIDIA H200 GPUs, while evaluation is performed on NVIDIA H20 GPUs.

\begin{figure*}[t]
  \includegraphics[width=\linewidth]{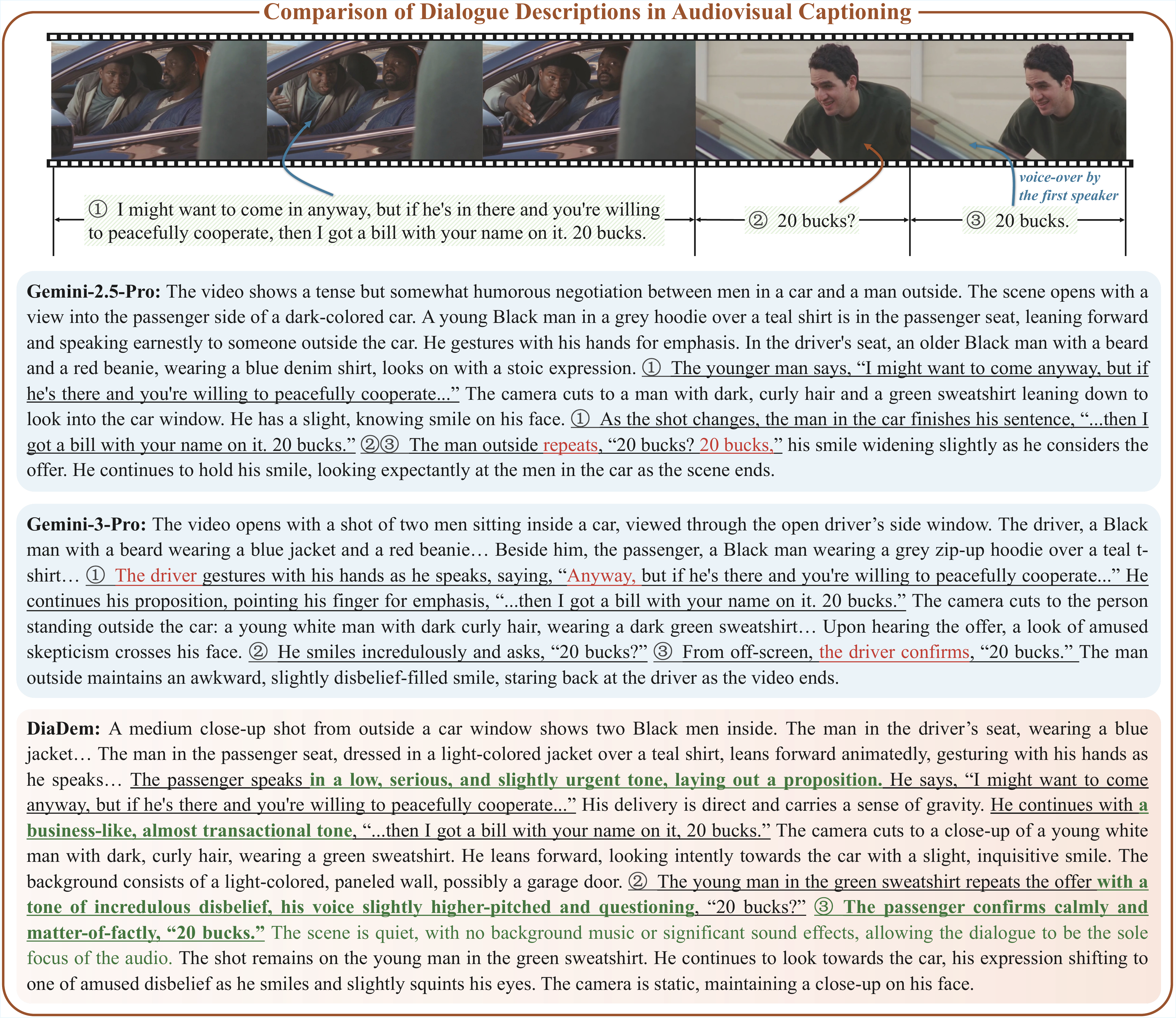}
  \caption{Qualitative comparison of DiaDem against two strong Gemini series models: Gemini-2.5-Pro and Gemini-3-Pro. Dialogue descriptions in the audiovisual captions are \underline{underlined}, and circled indices indicate correspondence with the respective ground-truth dialogues.}
  \label{fig:app_case_2}
\end{figure*}

\section{Additional Qualitative Results}\label{app:additional_cases}
Figs.~\ref{fig:app_case_1} and~\ref{fig:app_case_2} present additional qualitative comparisons of audiovisual captioning results among DiaDem and two strong Gemini series models, Gemini-2.5-Pro and Gemini-3-Pro.

In Fig.~\ref{fig:app_case_1}, both Gemini models fail to capture the initial Mandarin utterance spoken by the Asian male. Moreover, Gemini-2.5-Pro misidentifies the first segment of the Asian female’s English speech as Chinese, and erroneously attributes the latter part of her English utterance to the young doctor.

In Fig.~\ref{fig:app_case_2}, Gemini-2.5-Pro incorrectly interprets the final voice-over as a repetition of the young man’s earlier utterance. Gemini-3-Pro, on the other hand, omits the portion of the first utterance preceding the word ``anyway'' in the first utterance and fails to distinguish whether the speaker inside the car is the driver or the passenger.

In contrast, benefiting from effective data construction and training strategies, DiaDem not only produces more accurate and comprehensive dialogue descriptions, but also captures richer audiovisual details. In addition, DiaDem provides precise and nuanced characterizations of the speakers’ vocal emotions, highlighting its overall effectiveness.

\begin{figure*}
\begin{tcolorbox}[colback=white!95!gray, colframe=gray!50!black, rounded corners, title={Prompts for Gemini-2.5-Pro to produce initial dialogue descriptions}]
You are a highly skilled assistant specializing in extracting conversational dialogue from a video.\\

Your task is to carefully analyze the given video and accurately identify and extract all dialogue content within it.\\

Please directly output the dialogue in the following format without adding any other content. If no dialogue is present, state: "None."\\

Dialogue format:\\
Speaker A Description: "Dialogue from speaker A."\\
Speaker B Description: "Dialogue from speaker B."\\
Speaker A Description: "Further dialogue..."\\

The dialogue content should use the language exactly as heard in the audio. Do not translate it or perform any other transformations.\\

The description for each speaker (e.g., "Person in red dress") should be simplified for brevity. The key is to be concise and clearly distinguish between speakers (e.g., "Man in red shirt" is sufficient).
\end{tcolorbox}
\caption{Prompts for Gemini-2.5-Pro to produce initial dialogue descriptions.}
\label{fig:prompt_initial_dialogue}
\end{figure*}

\begin{figure*}
\begin{tcolorbox}[colback=white!95!gray, colframe=gray!50!black, rounded corners, title={Prompts for Gemini-3-Pro to correct speaker attribution in initial dialogue descriptions}]
You are a highly capable assistant specialized in refining speaker attribution in the original dialogue descriptions.\\

You will be provided with a video and a set of dialogue transcripts extracted from it.\\
The dialogue format is as follows:\\

Speaker A Description: "Utterance from Speaker A."\\
Speaker B Description: "Utterance from Speaker B."\\
Speaker A Description: "Another utterance..."\\

Your task is to correct the speaker attribution in the original dialogue transcripts.\\

The description for each speaker (e.g., "Person in red dress") should be simplified for brevity. The key is to be concise and clearly distinguish between speakers (e.g., "Man in red shirt" is sufficient).\\

Dialogue transcript:\\
\{\}
\end{tcolorbox}
\caption{Prompts for Gemini-3-Pro to correct speaker attribution in initial dialogue descriptions.}
\label{fig:prompt_refine_dialogue}
\end{figure*}

\begin{figure*}
\begin{tcolorbox}[colback=white!95!gray, colframe=gray!50!black, rounded corners, title={Prompts for Gemini-3-Pro to integrate the refined dialogue descriptions into audiovisual captions}]
You are a highly capable assistant specialized in correcting dialogue references and ASR-related information in audiovisual captions.\\

You will be given:\\
1) A high-quality audiovisual caption whose dialogue references and ASR transcript details may be inaccurate.\\
2) A dialogue description that contains more accurate dialogue references and ASR information.\\

Your task is to **use the accurate dialogue references and ASR information from the dialogue description to refine only the corresponding dialogue references and ASR transcript in the audiovisual caption**. Importantly, you should also repair and improve speaker descriptions, ensuring correctness with clear reference first, and then enhancing fluency and coherence. Ensure that each spoken line has a clear speaker attribution.\\
Do NOT modify any other parts of the audiovisual caption.\\

The audiovisual caption:\\
\{\}\\

The dialogue description:\\
\{\}
\end{tcolorbox}
\caption{Prompts for Gemini-3-Pro to integrate the refined dialogue descriptions into audiovisual captions.}
\label{fig:prompt_systhesis_caption}
\end{figure*}

\begin{figure*}
\begin{tcolorbox}[colback=white!95!gray, colframe=gray!50!black, rounded corners, title={List of prompts used to evaluate the audiovisual captioning quality}]
1. Provide a comprehensive description of all the content in the video, leaving out no details. Be sure to include as much of the audio information as possible, and ensure that your descriptions of the audio and video are closely aligned.\\
2. Thoroughly describe everything in the video, capturing every detail. Include as much information from the audio as possible, and ensure that the descriptions of both audio and video are well-coordinated.\\
3. Please describe all the information in the video without sparing every detail in it. As you describe, you should also describe as much of the information in the audio as possible, and pay attention to the synchronization between the audio and video descriptions.\\
4. Offer a detailed description of the video, making sure to include every detail. Also, incorporate as much information from the audio as you can, and ensure that your descriptions of the audio and video are in sync.\\
5. Describe every aspect of the video in full detail, covering all the information it contains. Additionally, include as much of the audio content as you can, and make sure your descriptions of the audio and video are synchronized.\\
6. Please provide a thorough description of all the content in the video, including every detail. As you describe, ensure that you also cover as much information from the audio as possible, and be mindful of the synchronization between the audio and video as you do so.\\
7. Give a detailed account of everything in the video, capturing all the specifics. While doing so, also include as much information from the audio as possible, ensuring that the descriptions of audio and video are well-synchronized.
\end{tcolorbox}
\caption{List of prompts used to evaluate the audiovisual captioning quality of models in the absence of an official prompt. During evaluation, prompts are randomly sampled from this list.}
\label{fig:prompt_list}
\end{figure*}

\begin{figure*}
\begin{tcolorbox}[colback=white!95!gray, colframe=gray!50!black, rounded corners, title={Prompts to extract dialogues in audiovisual captions}]
You are a highly skilled assistant specializing in extracting conversational dialogue from text.\\

Your task is to carefully analyze the given description of a video and accurately identify and extract all dialogue content within it.\\

Please directly output the dialogue in the following format without adding any other content. If no dialogue is present, state: "None."\\

Dialogue format:\\
Speaker A Description: "Dialogue from speaker A."\\
Speaker B Description: "Dialogue from speaker B."\\
Speaker A Description: "Further dialogue..."\\

Guidelines:\\
- The speaker descriptions (e.g., "person in red dress") must be derived from the video description.\\
- Ensure the descriptions are **accurate and distinguishable first**, and **concise** second.\\
- - The same speaker must always use exactly the same description, and different speakers must have different descriptions.\\
- Only include dialogue content; do not infer or add lines not explicitly present.\\

Video description:\\
\{\}
\end{tcolorbox}
\caption{Prompts to extract dialogues in audiovisual captions.}
\label{fig:prompt_extract_dialogue}
\end{figure*}

\begin{figure*}
\begin{tcolorbox}[colback=white!95!gray, colframe=gray!50!black, rounded corners, title={Prompts to identify speaker consistency}]
Given a video and several pairs of descriptive phrases about a certain subject, please help me determine whether the subjects in each pair refer to the same entity in the video.\\

For each pair of phrases, respond with 'Yes' or 'No', separated by a single space, without any extra characters. For example, if three pairs of phrases are provided, a valid response format would be: 'Yes No Yes'.\\

Descriptive phrases (each line contains a single pair):\\
\{\}
\end{tcolorbox}
\caption{Prompts to identify speaker consistency.}
\label{fig:prompt_speaker_identify}
\end{figure*}

\section{Details of Prompts}
In Figs.~\ref{fig:prompt_initial_dialogue} to~\ref{fig:prompt_systhesis_caption}, we present the prompts used in our audiovisual captioning data construction pipeline introduced in Sec.~\ref{sec:sft_pipeline}. The process begins with Gemini-2.5-Pro, which generates initial dialogue descriptions that faithfully capture spoken content but may contain inaccurate speaker attributions (Fig.~\ref{fig:prompt_initial_dialogue}). Subsequently, Gemini-3-Pro refines these attributions to ensure correct speaker assignment (Fig.~\ref{fig:prompt_refine_dialogue}). Finally, Gemini-3-Pro synthesizes the corrected dialogue descriptions with rich audiovisual captions produced by AVoCaDO (Fig.~\ref{fig:prompt_systhesis_caption}), yielding high-quality audiovisual captions that feature precise dialogue descriptions.

Fig.~\ref{fig:prompt_list} lists the set of prompts used for models that lack officially recommended instructions for audiovisual captioning. These prompts are randomly sampled to assess both general audiovisual captioning capabilities and the ability to accurately describe dialogues within the generated captions.

Figs.~\ref{fig:prompt_extract_dialogue} and~\ref{fig:prompt_speaker_identify} illustrate the prompts used, respectively, to extract dialogue descriptions from audiovisual captions and to assess the consistency between predicted speaker attributions and the ground-truth annotations, constituting essential components of DiaDemBench.

\end{document}